\relax
\documentclass[letterpaper]{article} 
\usepackage{xcolor} 
\usepackage{aaai21}  
\usepackage{times}  
\usepackage{helvet} 
\usepackage{courier}  
\usepackage[hyphens]{url}  
\usepackage{graphicx} 
\usepackage[switch]{lineno}  %
\usepackage{natbib}  
\usepackage{caption} 
\usepackage{svg}
\usepackage{amssymb}
\usepackage{multirow}
\usepackage{booktabs}
\title{Explaining Convolutional Neural Networks through Attribution-Based Input Sampling and Block-Wise Feature Aggregation}

\author{
     Sam Sattarzadeh,\textsuperscript{\rm 1} Mahesh Sudhakar,\textsuperscript{\rm 1} Anthony Lem,\textsuperscript{\rm 2} \\ Shervin Mehryar,\textsuperscript{\rm 1} K. N. Plataniotis,\textsuperscript{\rm 1}
     Jongseong Jang,\textsuperscript{\rm 3} Hyunwoo Kim,\textsuperscript{\rm 3} \\ Yeonjeong Jeong,\textsuperscript{\rm 3} Sangmin Lee,\textsuperscript{\rm 3} Kyunghoon Bae\textsuperscript{\rm 3}\\
 }
 \affiliations{

     \textsuperscript{\rm 1}The Edward S. Rogers Sr. Department of Electrical \& Computer Engineering, University of Toronto\\
     \textsuperscript{\rm 2}Division of Engineering Science, University of Toronto\\
     \textsuperscript{\rm 3}LG AI Research\\
     {sam.sattarzadeh, mahesh.sudhakar}@mail.utoronto.ca;
     {j.jang, hwkim}@lgresearch.ai

 }

\begin{document}
\maketitle

\begin{abstract}
As an emerging field in Machine Learning, Explainable AI (XAI) has been offering remarkable performance in interpreting the decisions made by Convolutional Neural Networks (CNNs). To achieve visual explanations for CNNs, methods based on class activation mapping and randomized input sampling have gained great popularity. However, the attribution methods based on these techniques provide low-resolution and blurry explanation maps that limit their explanation ability. To circumvent this issue, visualization based on various layers is sought. In this work, we collect visualization maps from multiple layers of the model based on an attribution-based input sampling technique and aggregate them to reach a fine-grained and complete explanation. We also propose a layer selection strategy that applies to the whole family of CNN-based models, based on which our extraction framework is applied to visualize the last layers of each convolutional block of the model. Moreover, we perform an empirical analysis of the efficacy of derived lower-level information to enhance the represented attributions. Comprehensive experiments conducted on shallow and deep models trained on natural and industrial datasets, using both ground-truth and model-truth based evaluation metrics validate our proposed algorithm by meeting or outperforming the state-of-the-art methods in terms of explanation ability and visual quality, demonstrating that our method shows stability regardless of the size of objects or instances to be explained.
\end{abstract}

\section{Introduction}
\label{section:intro}





Deep Neural models based on Convolutional Neural Networks (CNNs) have rendered inspiring breakthroughs in a wide variety of computer vision tasks. However, the lack of interpretability hurdles the understanding of decisions made by these models. This diminishes the trust consumers have for CNNs and limits the interactions between users and systems established based on such models. Explainable AI (XAI) attempts to interpret these cumbersome models \cite{Survey2018}. The offered interpretation ability has put XAI in the center of attention in various fields, especially where any single false prediction can cause severe consequences (e.g., healthcare) or where regulations force automotive decision-making systems to provide users with explanations (e.g., criminal justice) \cite{LiptonSurvey}.

This work particularly addresses the problem of visual explainability, which is a branch of post-hoc XAI. This field aims to visualize the behavior of models trained for image recognition tasks \cite{newsurvey}. The outcome of these methods is a heatmap in the same size as the input image named ``explanation map", representing the evidence leading the model to decide.


 \begin{figure}[t]
     \centering
     \includegraphics[width=0.9\linewidth]{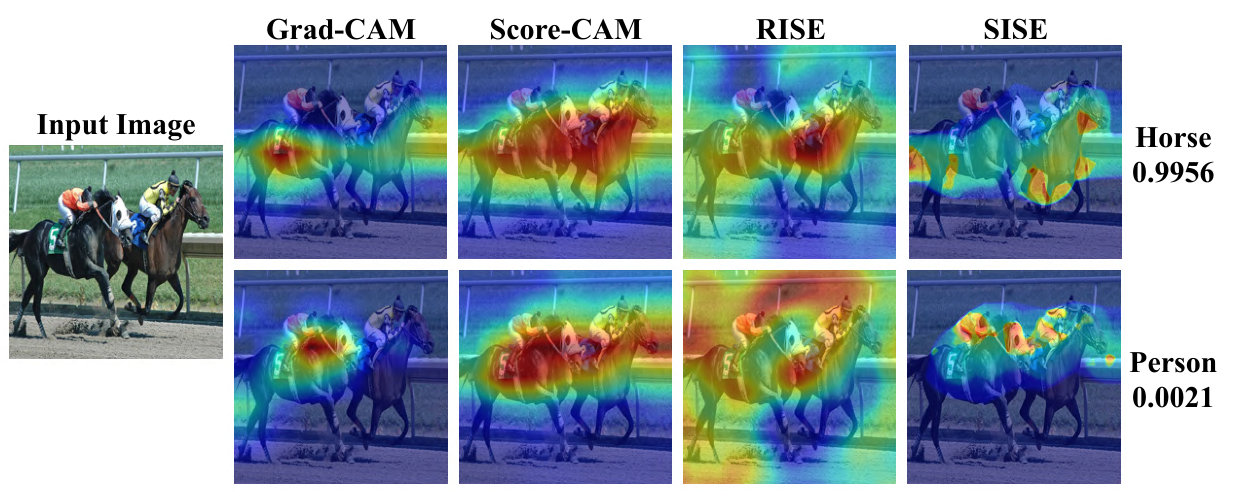}
     \caption{Comparison of conventional XAI methods with SISE (our proposed) to demonstrate SISE's ability to generate class discriminative explanations on a ResNet-50 model.}
     \label{fig:my_label}
 \end{figure}

Prior works on visual explainable AI can be broadly categorized into `approximation-based' \cite{Ribiero2016LIME}, `backpropagation-based', `perturbation-based', and `CAM-based' methodologies. In backpropagation-based methods, only the local attributions are represented, making them unable to measure global sensitivity. This drawback is addressed by image perturbation techniques used in recent works such as RISE \cite{petsiuk2018rise}, and Score-CAM \cite{wang2019score}. However, feedforwarding several perturbed images in these works makes them very slow. On the other hand, explanation maps produced by CAM-based methods suffer from a lack of spatial resolution as they are formed by combining the feature maps in the last convolutional layer of CNNs, which lack spatial information regarding the captured attributions.


In this work, we delve deeper into providing a solution for interpreting CNN-based models by analyzing multiple layers of the network. Our solution concentrates on mutual utilization of features represented inside a CNN in different semantic levels, achieving class discriminability and spatial resolution simultaneously. Inheriting productive ideas from the aforementioned types of approaches, we formulate a four-phase explanation method. In the first three phases, information extracted from multiple layers of the CNN is represented in their accompanying visualization maps. These maps are then combined via a fusion module to form a unique explanation map in the last phase. The main contributions of our work can be summarized as follows:

\begin{itemize}
    \item We introduce a novel XAI algorithm that offers both spatial resolution and explanation completeness in its output explanation map by 1) using multiple layers from the ``intermediate blocks'' of the target CNN, 2) selecting crucial feature maps from the outputs of the layers, 3) employing an attribution-based technique for input sampling to visualize the perspective of each layer, and 4) applying a feature aggregation step to reach refined explanation maps.
    \item We propose a strategy to select the minimum number of intermediate layers from a given CNN to probe and visualize their discovered features in order to provide the local explanations of the whole CNN. We discuss the applicability of this strategy to all of the feedforward CNNs.
    \item We conduct thorough experiments on various models trained on object detection and industrial anomaly classification datasets. To justify our method, we employ various metrics to compare our proposed method with other conventional approaches. Therefore, we show that the information between layers can be correctly combined to improve its inference's visual explainability.
\end{itemize}


\section{Related Work}

\subsubsection{Backpropagation-based methods} In general, calculating the gradient of a model's output to the input features or the hidden neurons is the basis of this type of explanation algorithms. The earliest backpropagation-based methods operate by computing the model's confidence score's sensitivity to each of the input features directly \cite{simonyan2013deep, zeiler2014visualizing}. To develop such methods, in some preceding works like DeepLift \cite{shrikumar2017learning}, IntegratedGradient \cite{sundararajan2017axiomatic} and SmoothGrad \cite{smilkov1706smoothgrad}, backpropagation-based equations are adapted to tackle the gradient issues. Some approaches such as LRP \cite{bach2015pixel}, SGLRP \cite{iwana2019explaining}, and RAP \cite{nam2020relative} modify backpropagation rules to measure the relevance or irrelevance of the input features to the model's prediction. Moreover, FullGrad \cite{srinivas2019full} and Excitation Backpropagation \cite{10.1007/s11263-017-1059-x} run by aggregating gradient information from several layers of the network.

\subsubsection{Perturbation-based methods}Several visual explanation methods probe the model's behavior using perturbed copies of the input. In general, various strategies can be chosen to perform input sampling. Like RISE \cite{petsiuk2018rise}, few of these approaches proposed random perturbation techniques to yield strong approximations of explanations. In Extremal Perturbation \cite{fong2019understanding}, an optimization problem is formulated to optimize a smooth perturbation mask maximizing the model's output confidence score. Most of the perturbation-based methods' noticeable property is that they treat the model like a ``black-box" instead of a ``white-box." 


\subsubsection{CAM-based methods}Based on the Class Activation Mapping method \cite{zhou2016learning}, an extensive research effort has been put to blend high-level features extracted by CNNs in a unique explanation map. CAM-based methods operate in three steps: 1) feeding the model with the input image, 2) scoring the feature maps in the last convolutional layer, and 3) combining the feature maps using the computed scores as weights. Grad-CAM \cite{selvaraju2017grad} and Grad-CAM++ \cite{chattopadhyay2017grad} utilize backpropagation in the second step which causes underestimation of sensitivity information due to gradient issues. Ablation-CAM  \cite{ramaswamy2020ablation}, Smooth Grad-CAM++ \cite{omeiza2019smooth}, and Score-CAM \cite{wang2019score}
have been developed to overcome these drawbacks.

Despite the strength of CAM-based methods in capturing the features extracted in CNNs, the lack of localization information in the coarse high-level feature maps limits such methods' performance by producing blurry explanations. Also, upsampling low-dimension feature maps to the size of input images distorts the location of captured features in some cases. Some recent works \cite{fanman2019meng, revisiting2020} addressed these limitations by amalgamating visualization maps obtained from multiple layers to achieve a fair trade-off between spatial resolution and class-distinctiveness of the features forming explanation maps.


\section{Methodology}

Our proposed algorithm is motivated by methods aiming to interpret the model's prediction using input sampling techniques. These methods have shown a great faithfulness in rationally inferring the predictions of models. However, they suffer from instability as their output depends on random sampling (RISE) or random initialization for optimizing a perturbation mask (Extremal perturbation). Also, such algorithms require an excessive runtime to provide their users with generalized results. To address these limitations, we advance a CNN-specific algorithm that improves their fidelity and plausibility (in the view of reasoning) with adaptive computational overhead for practical usage. We term our algorithm as Semantic Input Sampling for Explanation (SISE). To claim such a reform, we replace the randomized input sampling technique in RISE with a sampling technique that relies on the feature maps derived from multiple layers. We call this procedure \textit{attribution-based input sampling} and show that it provides the perspective of the model in various semantic levels, reducing the applicability of SISE to CNNs.



\begin{figure*}[t]
\centering
\includegraphics[width=0.8\linewidth]{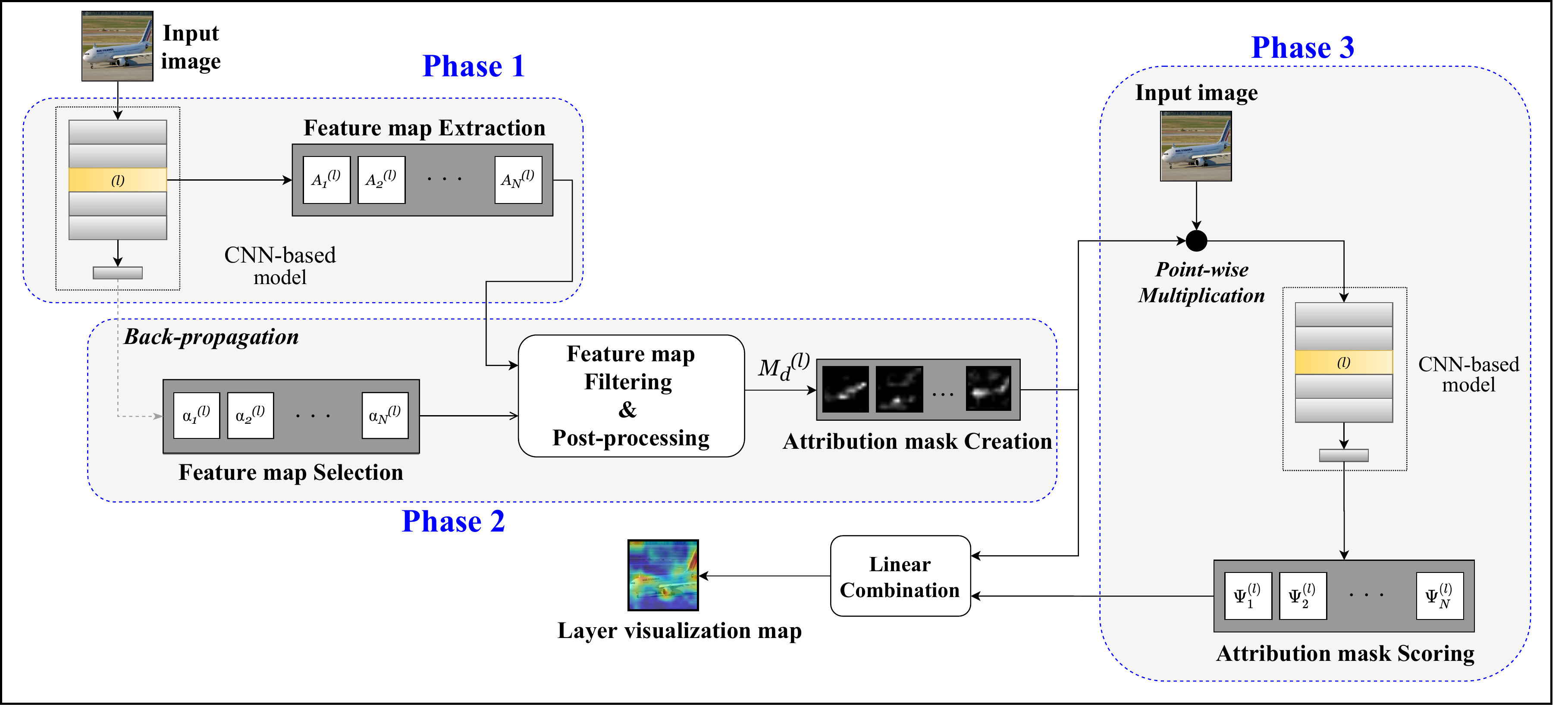}
\caption{Schematic of SISE's layer visualization framework (first three phases). The procedure in this framework is applied to multiple layers and is followed by the fusion framework (as in Fig. \ref{fig: polyfuse}).} 
\label{fig:SISE_layer_viz}
\end{figure*}

As sketched in Figs. \ref{fig:SISE_layer_viz} and \ref{fig: polyfuse}, SISE consists of four phases. In the first phase, multiple layers of the model are selected, and a set of corresponding output feature maps are extracted. For each set of feature maps in the second phase, a subset containing the most important feature maps is sampled with a backward pass. The selected feature maps are then post-processed to create sets of perturbation masks to be utilized in the third phase for attribution-based input sampling and are termed as \textit{attribution masks}. The first three phases are applied to multiple layers of the CNN to output a 2-dimensional saliency map named \textit{visualization map} for each layer. Such obtained visualization maps are aggregated in the last phase to reach the final explanation map.

In the following section, we present a \textit{block-wise} layer selection policy, showing that the richest knowledge in any CNN can be derived by probing the output of (the last layer in) each convolutional blocks, followed by the discussion of the phase-by-phase methodology of SISE.



\subsection{Block-Wise Feature Explanation }
\label{section:BWFE}



As we attempt to visualize multiple layers of the CNNs to merge spatial information and semantic information discovered by the CNN-based model, we intend to define the most crucial layers for explicating the model's decisions to reach a complete understanding of the model by visualizing the minimum number of layers.

Regardless of the specification of their architecture, all types of CNNs consist of convolutional blocks connected via pooling layers that aid the network to justify the existence of semantic instances. Each convolutional block is formed by cascading multiple layers, which may vary from a simple convolutional filter to more complex structures (e.g., bottleneck or MBConv layers). However, the dimensions of their input and output signal are the same. In a convolutional block, assuming the number of layers to be $L$, each $i$th layer can be represented with the function $f_{i}(.)$, where $i=\{1,...,L\}$. Denoting the input to each $i$th layer as $y_{i}$, the whole block can be mathematically described as $F(y_{1}) = f_{L}(y_{L})$. For plain CNNs (e.g., VGG, GoogleNet), the output of each convolutional block can be represented with the equation below:
\begin{equation}
    F(y_{1})=f_{L}(f_{L-1}(...(f_{1}(y_{1})))
    \label{eqls2}
\end{equation}

\begin{figure}[t]
 \centering
 \includegraphics[width=0.7\linewidth]{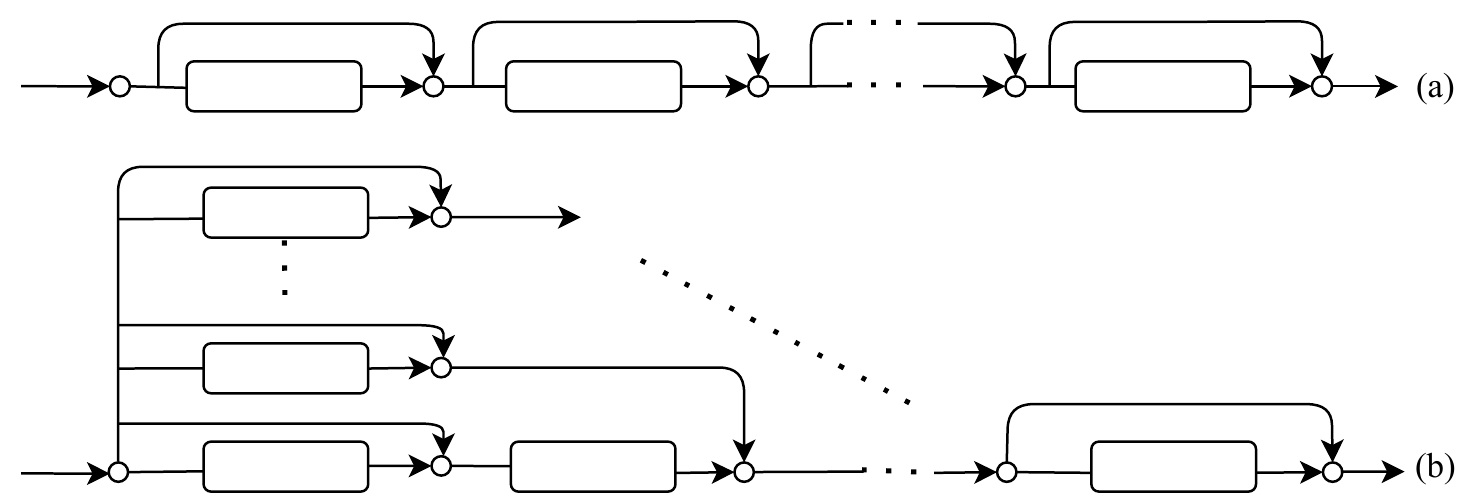}
 \caption{Architecture of the residual convolutional blocks as in \cite{shen2018end}. (a) raveled schematic of a residual network, (b) unraveled view of the residual network. }
 \label{unraveled}
\end{figure}

After the emergence of residual networks that utilize skip-connection layers to propagate the signals through a convolutional block in the families as ResNet models, DenseNet, EfficientNet \cite{EfficientNet, DenseNet, MobileNetv2}, and the models whose architecture are adaptively learned \cite{NAS}, it is debated that these neural networks can be represented with a more complicated view.  These types of networks can be viewed by the unraveled perspective, as presented in \cite{Veit}. Based on this perspective as in Fig. \ref{unraveled}, the connection between the input and output is formulated as follows:
\begin{equation}
    y_{i+1} = f_{i}(y_{i}) + y_{i}
    \label{eqls3}
\end{equation}
and hence,
\begin{equation}
    F(y_{1})=y_{1} + f_{1}(y_{1}) + ... + f_L(y_{1} + ... + f_{L-1}(y_{L-1}))
    \label{eqls4}
\end{equation}

The unraveled architecture as in Fig. \ref{unraveled} is comprehensive enough to be generalized even to shallower CNN-based models that lack skip-connection layers. For plain networks, the layer functions $f_{i}$ can be decomposed to an identity function $I$ and a residual function $g_{i}$ as follows:
\begin{equation}
    f_{i}(y_{i}) = I(y_{i}) + g_{i}(y_{i})
    \label{reform}
\end{equation}

Such a decomposition, yields to a similar equation form as equation \ref{eqls3}, and consequently, equation \ref{eqls4}.
\begin{equation}
    y_{i+1} = g_{i}(y_{i}) + y_{i}
    \label{eqls5}
\end{equation}

It can be inferred from the unraveled view that while feeding the model with an input, signals might not pass through all convolutional layers as they may skip some layers and be propagated to the next ones directly. However, this is not the case for pooling layers. Considering they change the signals' dimensions, equation \ref{reform} cannot be applied to such layers. To prove this hypothesis, an experiment was conducted in \cite{Veit}, where the corresponding test errors are reported for removing a layer individually from a residual network. It was observed that a significant degradation in test performance is recorded only when the pooling layers are removed.

Based on this hypothesis and result, most of the information in each model can be collected by probing the pooling layers. Thus, by visualizing these layers, it is possible to track the way features are propagated through convolutional blocks. Therefore, we derive attribution masks from the feature maps in the last layers of all of their convolutional blocks for any given CNN. Then, for each of these layers, we build a corresponding visualization map. These maps are utilized to perform a \textit{block-wise feature aggregation} in the last phase of our method.


\subsection{Feature Map Selection}
\label{subsection:phase1and2}

As discussed, the first two phases of SISE take responsibility to create multiple sets of attribution masks. In the first phase, we feed the model with an input image to derive sets of feature maps from various layers of the model. Then, we sample the most deterministic feature maps among each set and post-process them to obtain corresponding sets of attribution masks. These masks are utilized for performing attribution-based input sampling.

Assume $\Psi:\mathcal{I}\rightarrow\mathbb{R}$ be a trained model that outputs a confidence score for a given input image, where $\mathcal{I}$ is the space of RGB images $\mathcal{I}=\{I|I:\Lambda\rightarrow\mathbb{R}^{3}\}$, and $\Lambda=\{1,...,H\} \times \{1,...,W\}$ is the set of locations (pixels) in the image. Given any model and image, the goal of an explanation algorithm is to reach an explanation map $S_{I,\Psi}(\lambda)$, that assigns an ``importance value" to each location in the image $(\lambda \in \Lambda)$. Also, let $l$ be a layer containing $N$ feature maps represented as $A_{k}^{(l)} (k=\{1,...,N\})$ and the space of locations in these feature maps be denoted as $\Lambda^{(l)}$. These feature maps are collected by probing the feature extractor units of the model, and a similar strategy is also utilized in \cite{wang2019score}. The feature maps are formed in these units independently from the classifier part of the model. Thus, using the whole set of feature maps does not reflect the outlook of CNN's classifier.

\begin{figure}[htbp]
\centering
\includegraphics[width=0.85\linewidth]{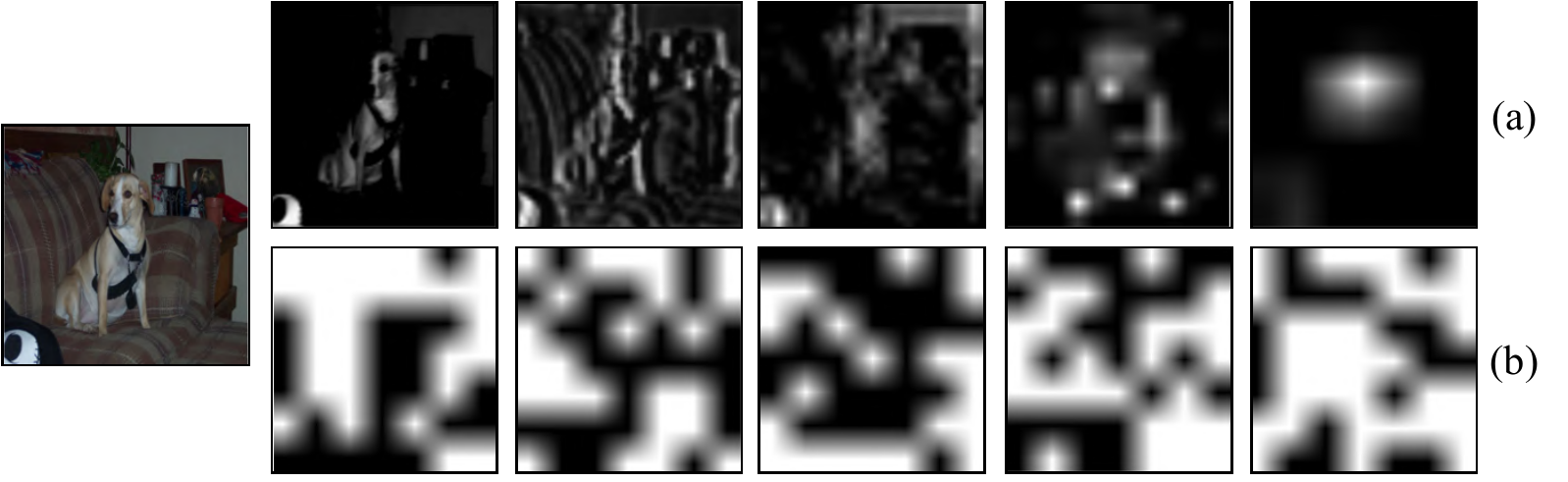}
\caption{Qualitative comparison of (a) attribution masks derived from different blocks of a VGG16 network as in SISE, with (b) random masks employed in RISE.}
\label{maskcomp}
\end{figure}

To identify and reject the class-indiscriminative feature maps, we partially backpropagate the signal to the layer $l$ to score the average gradient of model's confidence score to each of the feature maps. These average gradient scores are represented as follows:
\begin{equation}
    \alpha_k^{(l)} = \sum_{\lambda^{(l)} \in \Lambda^{(l)}} \frac{\partial \Psi(I)}{\partial A_{k}^{(l)} (\lambda^{(l)})}
    \label{Gradscore}
\end{equation}
The feature maps with corresponding non-positive average gradient scores - $\alpha_k^{(l)}$, tend to contain features related to other classes rather than the class of interest. Terming such feature maps as `negative-gradient', we define the set of attribution masks obtained from the `positive-gradient' feature maps, $M_d^{(l)}$, as:
\begin{equation}
    M_d^{(l)}=\{ \Omega(A_{k}^{(l)})|k\in \{1,...,N\}, \alpha_k^{(l)}>\mu \times \beta^{(l)} \}
    \label{maskeq}
\end{equation}
where $\beta^{(l)}$ denotes the maximum average gradient recorded.
\begin{equation}
    \beta^{(l)} = \max_{k\in \{1,...,N\}} ( \alpha_k^{(l)})
\end{equation}
In equation \ref{maskeq}, $\mu \in \mathbb{R}_{\geq 0}$ is a threshold parameter that is 0 by default to discard negative-gradient feature maps while retaining only the positive-gradients. Furthermore, $\Omega(.)$ represents a post-processing function that converts feature maps to attribution masks. This function contains a `bilinear interpolation,' upsampling the feature maps to the size of the input image, followed by a linear transformation that normalizes the values in the mask in the range $[0,1]$. A visual comparison of attribution masks and random masks in Fig. \ref{maskcomp} emphasizes such advantages of the former.

\subsection{Attribution-Based Input Sampling}
\label{subsection:phase3}


Considering the same notations as the previous section, and according to RISE method, the confidence scores observed for the copies of an image masked with a set of binary masks $(M: \Lambda \rightarrow \{0,1\})$ are used to form the explanation map by,
\begin{equation}
    S_{I,\Psi}(\lambda)={\mathbb{E}}_{M}[\Psi(I \odot m)|m(\lambda)=1]
    \label{eq01}
\end{equation}
where $I \odot m\ $ denotes a masked image obtained by point-wise multiplication between the input image and a mask $m\in M$.
The representation of equation \ref{eq01} can be modified to be generalized for sets of smooth masks $(M: \Lambda \rightarrow [0,1])$. Hence, we reformat equation \ref{eq01} as:
\begin{equation}
    S_{I,\Psi}(\lambda)=\mathbb{E}_{M} [\Psi(I\odot m)\cdot C_{m}(\lambda)] 
    \label{SISE_LV}
\end{equation}
where the term $C_{m}(\lambda)$ indicates the contribution amount of each pixel in the masked image. Setting the contribution indicator as $C_{m}(\lambda)=m(\lambda)$, makes equation \ref{SISE_LV} equivalent to equation \ref{eq01}. We normalize these scores according to the size of perturbation masks to decrease the assigned reward to the background pixels when a high score is reached for a mask with too many activated pixels. Thus, we define this term as:
\begin{equation}
    C_{m}(\lambda)=\frac{m(\lambda)}{\sum_{\lambda\in\Lambda}m(\lambda)}
    \label{contribution}
\end{equation}
Such a formulation increases the concentration on smaller features, particularly when multiple objects (either from the same instance or different ones) are present in an image.

Putting block-wise layer selection policy and attribution mask selection strategy together with the modified RISE framework, for each CNN containing $B$ convolutional blocks, the last layer of each block is indicated as $l_{b}\in\{1,..., B\}$. Using equations \ref{SISE_LV} and \ref{contribution}, we form corresponding visualization maps for each of these layers by:
\begin{equation}
    V_{I,\Psi}^{(l_{b})}(\lambda)=\mathbb{E}_{M_d^{(l_{b})}} [\Psi(I\odot m)\cdot C_{m}(\lambda)] 
    \label{VL_all}
\end{equation}

\subsection{Fusion Module}
In the fourth phase of SISE, the flow of features from low-level to high-level blocks are tracked. The inputs to the fusion module are the visualization layers obtained from the third phase of SISE. On the other hand, this module's output is a 2-dimensional explanation map, which is the output of SISE. The fusion block is responsible for correcting spatial distortions caused by upsampling coarse feature maps to higher dimensions and refining the localization of attributions derived from the model.
\begin{figure}[htbp]
\centering
\includegraphics[width=0.9\linewidth]{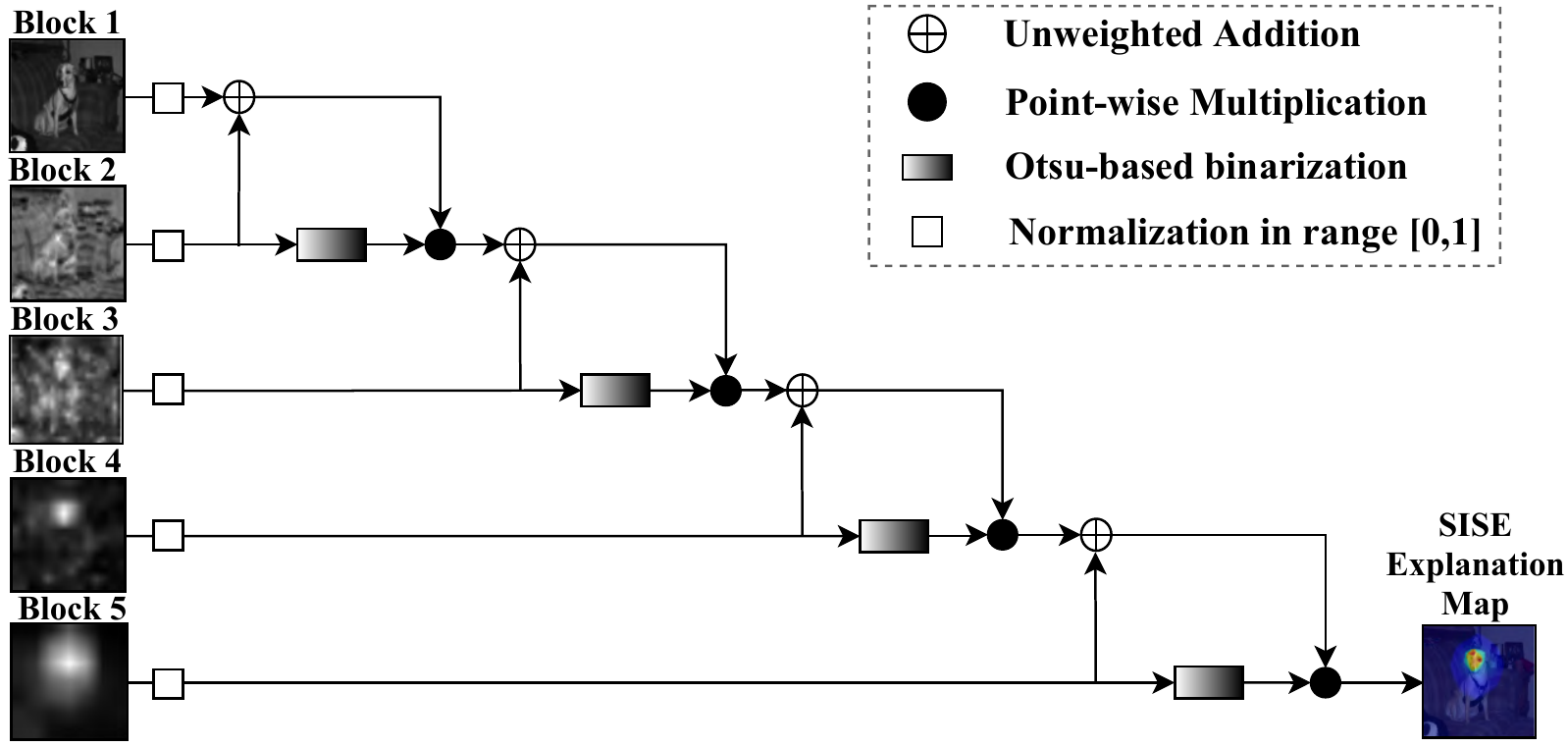}
\caption{SISE fusion module for a CNN with 5 convolutional blocks.}
\label{fig: polyfuse}
\end{figure}

Our fusion module is designed with cascaded fusion blocks. In each block, the feature information from the visualization maps representing explanations for two consecutive blocks is collected using an ``addition" block. Then, the features that are absent in the latter visualization map are removed from the collective information by masking the output of the addition block with a binary mask indicating the activated regions in the latter visualization map. To reach the binary mask, we apply an adaptive threshold to the latter visualization map, determined by Otsu's method \cite{Otsu}. By cascading fusion blocks as in Fig. \ref{fig: polyfuse}, the features determining the model's prediction are represented in a more fine-grained manner while the inexplicit features are discarded.

\begin{figure*}[t]
\centering
\includegraphics[width=0.85\linewidth]{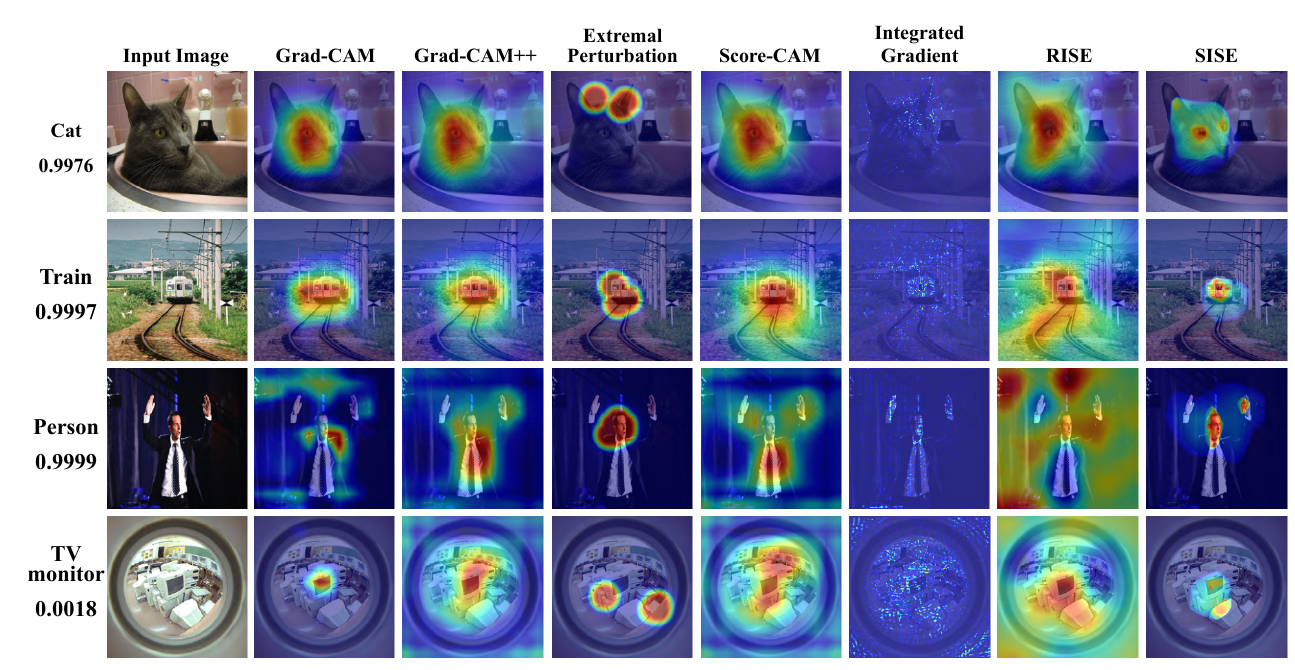}
\caption{Qualitative comparison of the state-of-the-art XAI methods with our proposed SISE for test images from the PASCAL VOC 2007 dataset. The first two rows are the results from a ResNet-50 model, and the last two are from a VGG16 model.} 
\label{fig:voc_qual_results}
\end{figure*}

\section{Experiments}
\label{section:exps}


We verify our method's performance on shallow and deep CNNs, including VGG16, ResNet-50, and ResNet-101 architectures. To conduct the experiments, we employed PASCAL VOC 2007 \cite{PASCALVOC} and Severstal \cite{severstalDataset} datasets. The former is a popular object detection dataset containing 4952 test images belonging to 20 object classes. As images with many small object occurrences and multiple instances of different classes are prevalent in this dataset, it is hard for an XAI algorithm to perform well on the whole dataset. The latter is an industrial steel defect detection dataset created for anomaly detection and steel defect segmentation problems. We reformatted it into a defect classification dataset instead, containing 11505 test images from 5 different classes, including one normal class and four different defects classes. Class imbalance, intraclass variation, and interclass similarity are the main challenges of this recast dataset.

\subsection{Experimental Setup}

Experiments conducted on the PASCAL VOC 2007 dataset are evaluated on its test set with a VGG16, and a ResNet-50 model from the TorchRay library \cite{fong2019understanding}, trained by \cite{10.1007/s11263-017-1059-x}, both trained for multi-label image classification. The top-5 accuracies of the models on the test set are 93.29\% and 93.09\%, respectively. On the other hand, for conducting experiments on Severstal, we trained a ResNet-101 model (with a test accuracy of 86.58\%) on the recast dataset to assess the performance of the proposed method in the task of visual defect inspection. To recast the Severstal dataset for classification, the train and test images were cropped into patches of size $256 \times 256$. In our evaluations, a balanced subset of 1381 test images belonging to defect classes labeled as 1, 2, 3, and 4 is chosen. We have implemented SISE on Keras and set the parameter $\mu$ to its default value, 0.

\subsection{Qualitative Results}
Based on explanation quality, we have compared SISE with other state-of-the-art methods on sample images from the Pascal dataset in Fig. \ref{fig:voc_qual_results} and Severstal dataset in Fig. \ref{svs_qual}. Images with both normal-sized and small object instances are shown along with their corresponding confidence scores. Moreover, Figs. \ref{fig:my_label} and \ref{twoclass_sise} with images of multiple objects from different classes depict the superior ability of SISE in discriminating the explanations of various classes in comparison with other methods and RISE in particular.



\subsection{Quantitative Results}
Quantitative analysis includes evaluation results categorized into `ground truth-based' and `model truth-based' metrics. The former is used to justify the model by assessing the extent to which the algorithm satisfies the users by providing visually superior explanations, while the latter is used to analyze the model behavior by assessing the faithfulness of the algorithm and its correctness in capturing the attributions in line with the model's prediction procedure. The reported results of RISE and Extremal Perturbation in Table \ref{tab: gt_metrics} are averaged on three runs. The utilized metrics are discussed below.

\subsubsection{Ground truth-based Metrics:}
The state-of-the-art explanation algorithms are compared with SISE based on three distinct ground-truth based metrics to justify the visual quality of the explanation maps generated by our method. Denoting the ground-truth mask as $G$ and the achieved explanation map as $S$, the evaluation metrics used are:
\begin{figure}[t]
\centering
\includegraphics[width=0.9\linewidth]{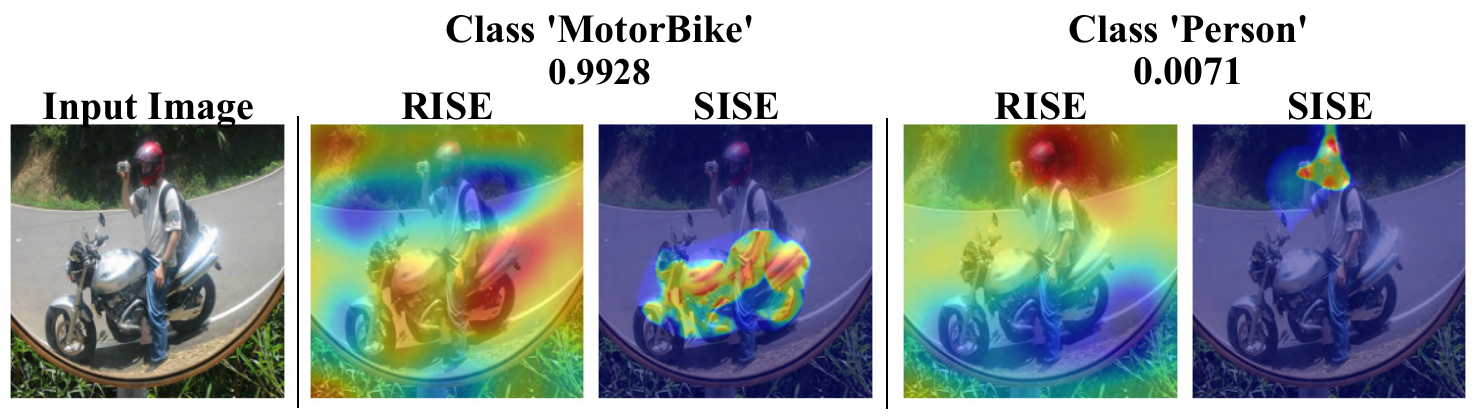}
\caption{Class discriminative ability of SISE vs. RISE obtained from a VGG16 model}
\label{twoclass_sise}
\end{figure}

\textbf{Energy-Based Pointing Game (EBPG)} evaluates the precision and denoising ability of XAI algorithms \cite{wang2019score}. Extending the traditional Pointing Game, EBPG considers all pixels in the resultant explanation map $S$ for evaluation by measuring the fraction of its energy captured in the corresponding ground truth $G$, as $EBPG = \frac{||S \odot G||_{1}}{||S||_{1}}$.

\textbf{mIoU} analyses the localization ability and meaningfulness of the attributions captured in an explanation map. In our experiments, we select the top 20\% pixels highlighted in each explanation map $S$ and compute the mean intersection over union with their corresponding ground-truth masks.

\begin{table*}[t]
 \centering
 \begin{tabular}{c c c c c c c c c c}
 \toprule
 \multirow{2}{*}{\textbf{Model}} & \multirow{2}{*}{\textbf{Metric}} &
 \multirow{2}{*}{Grad-CAM} & Grad- & Extremal & \multirow{2}{*}{RISE} & Score- & Integrated & \multirow{2}{*}{FullGrad} & \multirow{2}{*}{\textbf{SISE}} \\
 & & & CAM++ & Perturbation & & CAM & Gradient & & \\
 \midrule
 \multirow{5}{*}{\textbf{VGG16}} & \textbf{EBPG} &
 55.44 & 46.29 & \textbf{61.19} & 33.44 & 46.42 & 36.87 & 38.72 & \underline{60.54} \\
  & \textbf{mIoU} &
 26.52 & \textbf{28.1} & 25.44 & 27.11 & 27.71 & 14.11 & 26.61 & \underline{27.79} \\
  & \textbf{Bbox} &
 51.7 & \underline{55.59} & 51.2 & 54.59 & 54.98 & 33.97 & 54.17 & \textbf{55.68} \\
 \cmidrule(lr){2-10}
 & \textbf{Drop} & 49.47 & 60.63 & 43.90 & \underline{39.62} & 39.79 & 64.74 & 60.78 & \textbf{38.40} \\
  & \textbf{Increase} & 31.08 & 23.89 & 32.65 & \underline{37.76} & 36.42 & 26.17 & 22.73 & \textbf{37.96} \\
 \midrule
 \multirow{5}{*}{\textbf{ResNet-50}} & \textbf{EBPG} &
60.08 & 47.78 & \underline{63.24} & 32.86 & 35.56 & 40.62 & 39.55 & \textbf{66.08} \\
  & \textbf{mIoU} &
 \textbf{32.16} & 30.16 & 26.29 & 27.4 & 31.0 & 15.41 & 20.2& \underline{31.37} \\
  & \textbf{Bbox} &
 \underline{60.25} & 58.66 & 52.34 & 55.55 & 60.02 & 34.79 & 44.94 & \textbf{61.59} \\
  \cmidrule(lr){2-10}
  & \textbf{Drop} &
 35.80 & 41.77 & 39.38 & 39.77 & \underline{35.36} & 66.12 & 65.99 & \textbf{30.92} \\
  & \textbf{Increase} &
 36.58 & 32.15 & 34.27 & \underline{37.08} & \underline{37.08} & 24.24 & 25.36 & \textbf{40.22} \\
 \bottomrule
 \end{tabular}
 \caption{Results of ground truth-based and model truth-based metrics for state-of-the-art XAI methods along with SISE (proposed) on two networks trained on the PASCAL VOC 2007 dataset. For each metric, the best is shown in bold, and the second-best is underlined.  Except for Drop\%, the higher is better for all other metrics. All values are reported in percentage.}
 \label{tab: gt_metrics}
\end{table*}
\textbf{Bounding box (Bbox)}  \cite{Schulz2020Restricting} is taken into account as a size-adaptive variant of mIoU. Considering $N$ as the number of ground truth pixels in $G$, the Bbox score is calculated by selecting the top $N$ pixels in $S$ and evaluating the corresponding fraction captured over $G$.
\begin{figure}[t]
\centering
\includegraphics[width=0.9\linewidth]{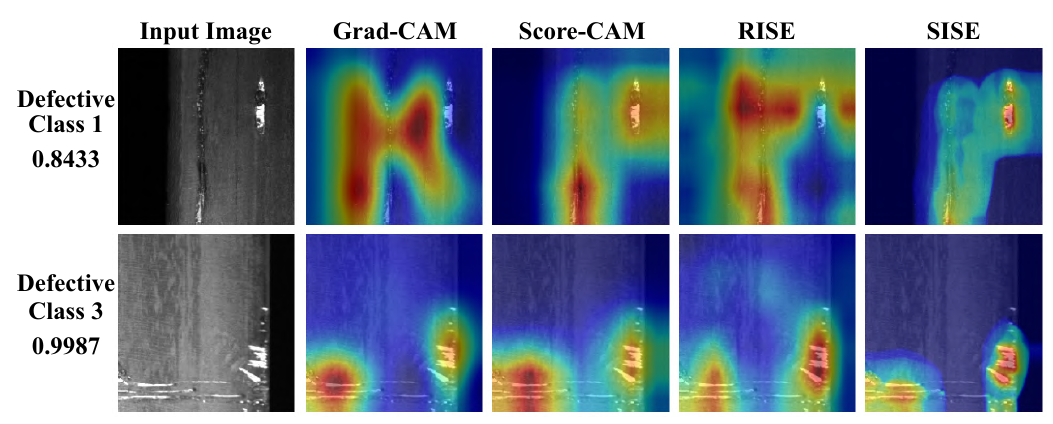}
\caption{Qualitative comparison of explanation maps by a ResNet-101 model on test images from Severstal dataset.}
\label{svs_qual}
\end{figure}
\subsubsection{Model truth-based metrics:}
To evaluate the correlation between the representations of our method and the model's predictions, model-truth based metrics are employed to compare SISE with the other state-of-the-art methods. As visual explanation algorithms' main objective is to envision the model's perspective for its predictions, these metrics are considered of higher importance.

\textbf{Drop\% and Increase\%}, as introduced in \cite{chattopadhyay2017grad} and later modified by \cite{ramaswamy2020ablation, fu2020axiombased}, can be interpreted as an indicator of the positive attributions missed and the negative attribution discarded from the explanation map respectively. Given a model $\Psi(.)$, an input image $I_{i}$ from a dataset containing $K$ images, and an explanation map $S(I_{i})$, the Drop/Increase \% metric selects the most important pixels in $S(I_{i})$ to measure their contribution towards the model's prediction. A threshold function $T(.)$ is applied on $S(I_{i})$ to select the top 15\% pixels that are then extracted from $I_{i}$ using point-wise multiplication and fed to the model. The confidence scores on such perturbed images are then compared with the original score, according to the equations $Drop\%=\frac{1}{K}\sum_{i=1}^{K}\frac{\max(0,\Psi(I_{i})-\Psi(I_{i}\odot T(I_{i})))}{\Psi(I_{i})}\times100$ and $Increase\%=\sum_{i=1}^{K} sign(\Psi(I_{i}\odot T(I_{i}))-\Psi(I_{i}))$.

\begin{table}[th]
 \centering
 \begin{tabular}{c c c}
 \toprule
 \textbf{XAI method} & \textbf{Drop\%} & \textbf{Increase\%}\\
 \midrule
 Grad-CAM & 67.44 & 12.46\\
 Grad-CAM++ & 64.1 & 12.96\\
 RISE & \underline{63.25} & \underline{15.63}\\
 Score-CAM & 64.29 & 10.35\\
 FullGrad & 77.23 & 10.26\\
 \textbf{SISE} &\textbf{61.06} & \textbf{15.64}\\
 \bottomrule
 \end{tabular}
 \caption{Results of model truth-based metrics of SISE and state-of-the-art algorithms on a ResNet-101 model trained on Severstal data set.}
 \label{tab: sev_metrics}
\end{table}

\section{Discussion}

The experimental results in Figs. \ref{fig:my_label},  \ref{fig:voc_qual_results}, \ref{twoclass_sise}, and \ref{svs_qual}  demonstrate the resolution, and concreteness of SISE explanation maps, which is further supported by justifying our method via ground truth-based evaluation metrics as in Table \ref{tab: gt_metrics}. Also, model truth-based metrics in Tables \ref{tab: gt_metrics} and \ref{tab: sev_metrics} prove SISE's supremacy in highlighting the evidence, based on which the model makes a prediction. Similar to the CAM-based methods, the output of the last convolutional block plays the most critical role in our method. However, by considering the intermediate layers based on the block-wise layer selection, SISE's advantageous properties are enhanced. Furthermore, utilizing attribution-based input sampling instead of a randomized sampling, ignoring the unrelated feature maps, and modifying the linear combination step dramatically improves the visual clarity and completeness offered by SISE.

\subsubsection{Complexity Evaluation}
In addition to performance evaluations, a runtime test is carried out to compare the complexity of the methods, using a Tesla T4 GPU with 16GB of memory and the ResNet-50 model. Reported runtimes were averaged over 100 trials using random images from the PASCAL VOC 2007 test set. Grad-CAM and Grad-CAM++ achieved the best runtimes, 19 and 20 milliseconds, respectively. On the other hand, Extremal Perturbation recorded the longest runtime, 78.37 seconds, since it optimizes numerous variables. In comparison with RISE, which has a runtime of 26.08 seconds, SISE runs in 9.21 seconds.

\subsubsection{Ablation Study}
While RISE uses around 8000 random masks to operate on a ResNet-50 model, SISE uses around 1900 attribution masks with $\mu$ set to 0, out of a total of 3904 feature maps initially extracted from the same ResNet-50 model before negative-gradient feature maps were removed. The difference in the number of masks allows SISE to operate in around 9.21 seconds. To analyze the effect of reducing the number of attribution masks on SISE's performance, an ablation study is carried. By changing $\mu$ to 0.3, a scanty variation in the boundary of explanation maps can be noticed while the runtime is reduced to 2.18 seconds. This shows that ignoring feature maps with low gradient values does not considerably affect SISE outputs since they tend to be assigned low scores in the third phase of SISE anyway. By increasing $\mu$ to 0.5, a slight decline in the performance is recorded along with a runtime of just 0.65 seconds. 

A more detailed analysis of the effect of $\mu$ on various evaluation metrics along with an extensive discussion of our algorithm and additional results on MS COCO 2014 dataset \cite{lin2014microsoft} are provided in the technical appendix of our extended version on arXiv\footnote{\url{https://arxiv.org/abs/2010.00672}}.

\section{Conclusion}
\label{section:future}
In this work, we propose SISE - a novel visual explanation algorithm specialized to the family of CNN-based models. SISE generates explanations by aggregating visualization maps obtained from the output of convolutional blocks through attribution-based input sampling. Qualitative results show that our method can output high-resolution explanation maps, the quality of which is emphasized by quantitative analysis using ground truth-based metrics. Moreover, model truth-based metrics demonstrate that our method also outperforms other state-of-the-art methods in providing concrete explanations. Our experiments reveal that mutual utilization of features captured in final and intermediate layers of the model aids in producing explanation maps that accurately locate object instances and reach a greater portion of attributions leading the model to make a decision.

\section{Acknowledgement}

This research was supported by LG AI Research. The authors thank all anonymous reviewers for their detailed suggestions and critical comments on the original manuscript that substantially helped to improve the clarity of this paper.

\bibliography{bibliography}

\begin{thebibliography}{37}
\providecommand{\natexlab}[1]{#1}
\providecommand{\url}[1]{\texttt{#1}}
\providecommand{\urlprefix}{URL }
\expandafter\ifx\csname urlstyle\endcsname\relax
  \providecommand{\doi}[1]{doi:\discretionary{}{}{}#1}\else
  \providecommand{\doi}{doi:\discretionary{}{}{}\begingroup
  \urlstyle{rm}\Url}\fi

\bibitem[{Adebayo et~al.(2018)Adebayo, Gilmer, Muelly, Goodfellow, Hardt, and
  Kim}]{KimSanity}
Adebayo, J.; Gilmer, J.; Muelly, M.; Goodfellow, I.; Hardt, M.; and Kim, B.
  2018.
\newblock Sanity Checks for Saliency Maps.
\newblock In Bengio, S.; Wallach, H.; Larochelle, H.; Grauman, K.;
  Cesa-Bianchi, N.; and Garnett, R., eds., \emph{Advances in Neural Information
  Processing Systems}, volume~31, 9505--9515. Curran Associates, Inc.
\newblock
  \urlprefix\url{https://proceedings.neurips.cc/paper/2018/file/294a8ed24b1ad22ec2e7efea049b8737-Paper.pdf}.

\bibitem[{Bach et~al.(2015)Bach, Binder, Montavon, Klauschen, M{\"u}ller, and
  Samek}]{bach2015pixel}
Bach, S.; Binder, A.; Montavon, G.; Klauschen, F.; M{\"u}ller, K.-R.; and
  Samek, W. 2015.
\newblock On pixel-wise explanations for non-linear classifier decisions by
  layer-wise relevance propagation.
\newblock \emph{PloS one} 10(7): e0130140.

\bibitem[{Barredo~Arrieta et~al.(2019)Barredo~Arrieta, Diaz~Rodriguez, Del~Ser,
  Bennetot, Tabik, Barbado~González, Garcia, Gil-Lopez, Molina, Benjamins,
  Chatila, and Herrera}]{newsurvey}
Barredo~Arrieta, A.; Diaz~Rodriguez, N.; Del~Ser, J.; Bennetot, A.; Tabik, S.;
  Barbado~González, A.; Garcia, S.; Gil-Lopez, S.; Molina, D.; Benjamins,
  V.~R.; Chatila, R.; and Herrera, F. 2019.
\newblock Explainable Artificial Intelligence (XAI): Concepts, Taxonomies,
  Opportunities and Challenges toward Responsible AI.
\newblock \emph{Information Fusion} \doi{10.1016/j.inffus.2019.12.012}.

\bibitem[{{Chattopadhay} et~al.(2018){Chattopadhay}, {Sarkar}, {Howlader}, and
  {Balasubramanian}}]{chattopadhyay2017grad}
{Chattopadhay}, A.; {Sarkar}, A.; {Howlader}, P.; and {Balasubramanian}, V.~N.
  2018.
\newblock Grad-CAM++: Generalized Gradient-Based Visual Explanations for Deep
  Convolutional Networks.
\newblock In \emph{2018 IEEE Winter Conference on Applications of Computer
  Vision (WACV)}, 839--847.
\newblock \doi{10.1109/WACV.2018.00097}.

\bibitem[{Everingham et~al.(2007)Everingham, Van~Gool, Williams, Winn, and
  Zisserman}]{PASCALVOC}
Everingham, M.; Van~Gool, L.; Williams, C. K.~I.; Winn, J.; and Zisserman, A.
  2007.
\newblock The {PASCAL} {V}isual {O}bject {C}lasses {C}hallenge 2007 {(VOC2007)}
  {R}esults.
\newblock
  \urlprefix\url{http://www.pascal-network.org/challenges/VOC/voc2007/workshop/index.html}.

\bibitem[{Fong, Patrick, and Vedaldi(2019)}]{fong2019understanding}
Fong, R.; Patrick, M.; and Vedaldi, A. 2019.
\newblock Understanding deep networks via extremal perturbations and smooth
  masks.
\newblock In \emph{Proceedings of the IEEE International Conference on Computer
  Vision}, 2950--2958.

\bibitem[{Fu et~al.(2020)Fu, Hu, Dong, Guo, Gao, and Li}]{fu2020axiombased}
Fu, R.; Hu, Q.; Dong, X.; Guo, Y.; Gao, Y.; and Li, B. 2020.
\newblock Axiom-based Grad-CAM: Towards Accurate Visualization and Explanation
  of CNNs.
\newblock In \emph{British Machine Vision Conference}.

\bibitem[{Hoffman et~al.(2018)Hoffman, Mueller, Klein, and Litman}]{Survey2018}
Hoffman, R.~R.; Mueller, S.~T.; Klein, G.; and Litman, J. 2018.
\newblock Metrics for Explainable {AI:} Challenges and Prospects.
\newblock \emph{CoRR} abs/1812.04608.
\newblock \urlprefix\url{http://arxiv.org/abs/1812.04608}.

\bibitem[{Huang et~al.(2017)Huang, Liu, Van Der~Maaten, and
  Weinberger}]{DenseNet}
Huang, G.; Liu, Z.; Van Der~Maaten, L.; and Weinberger, K.~Q. 2017.
\newblock Densely connected convolutional networks.
\newblock In \emph{Proceedings of the IEEE conference on computer vision and
  pattern recognition}, 4700--4708.

\bibitem[{Iwana, Kuroki, and Uchida(2019)}]{iwana2019explaining}
Iwana, B.~K.; Kuroki, R.; and Uchida, S. 2019.
\newblock Explaining convolutional neural networks using softmax gradient
  layer-wise relevance propagation.
\newblock In \emph{2019 IEEE/CVF International Conference on Computer Vision
  Workshop (ICCVW)}, 4176--4185. IEEE.

\bibitem[{Lin et~al.(2014)Lin, Maire, Belongie, Hays, Perona, Ramanan,
  Doll{\'a}r, and Zitnick}]{lin2014microsoft}
Lin, T.-Y.; Maire, M.; Belongie, S.; Hays, J.; Perona, P.; Ramanan, D.;
  Doll{\'a}r, P.; and Zitnick, C.~L. 2014.
\newblock Microsoft coco: Common objects in context.
\newblock In \emph{European conference on computer vision}, 740--755. Springer.

\bibitem[{Lipton(2018)}]{LiptonSurvey}
Lipton, Z.~C. 2018.
\newblock The Mythos of Model Interpretability: In Machine Learning, the
  Concept of Interpretability is Both Important and Slippery.
\newblock \emph{Queue} 16(3): 31–57.
\newblock ISSN 1542-7730.
\newblock \doi{10.1145/3236386.3241340}.
\newblock \urlprefix\url{https://doi.org/10.1145/3236386.3241340}.

\bibitem[{Meng et~al.(2019)Meng, Huang, Li, and Wu}]{fanman2019meng}
Meng, F.; Huang, K.; Li, H.; and Wu, Q. 2019.
\newblock Class Activation Map Generation by Representative Class Selection and
  Multi-Layer Feature Fusion.
\newblock \emph{arXiv preprint arXiv:1901.07683} .

\bibitem[{Nam et~al.(2020)Nam, Gur, Choi, Wolf, and Lee}]{nam2020relative}
Nam, W.-J.; Gur, S.; Choi, J.; Wolf, L.; and Lee, S.-W. 2020.
\newblock Relative Attributing Propagation: Interpreting the Comparative
  Contributions of Individual Units in Deep Neural Networks.
\newblock In \emph{AAAI}, 2501--2508.

\bibitem[{Omeiza et~al.(2019)Omeiza, Speakman, Cintas, and
  Weldermariam}]{omeiza2019smooth}
Omeiza, D.; Speakman, S.; Cintas, C.; and Weldermariam, K. 2019.
\newblock Smooth grad-cam++: An enhanced inference level visualization
  technique for deep convolutional neural network models.
\newblock \emph{arXiv preprint arXiv:1908.01224} .

\bibitem[{{Otsu}(1979)}]{Otsu}
{Otsu}, N. 1979.
\newblock A Threshold Selection Method from Gray-Level Histograms.
\newblock \emph{IEEE Transactions on Systems, Man, and Cybernetics} 9(1):
  62--66.

\bibitem[{{PAO Severstal}(2019)}]{severstalDataset}
{PAO Severstal}. 2019.
\newblock Severstal: Steel Defect Detection on Kaggle Challenge.
\newblock
  \urlprefix\url{https://www.kaggle.com/c/severstal-steel-defect-detection}.

\bibitem[{Petsiuk, Das, and Saenko(2018)}]{petsiuk2018rise}
Petsiuk, V.; Das, A.; and Saenko, K. 2018.
\newblock RISE: Randomized Input Sampling for Explanation of Black-box Models.
\newblock In \emph{Proceedings of the British Machine Vision Conference
  (BMVC)}.

\bibitem[{Ramaswamy et~al.(2020)}]{ramaswamy2020ablation}
Ramaswamy, H.~G.; et~al. 2020.
\newblock Ablation-CAM: Visual Explanations for Deep Convolutional Network via
  Gradient-free Localization.
\newblock In \emph{The IEEE Winter Conference on Applications of Computer
  Vision}, 983--991.

\bibitem[{Rebuffi et~al.(2020)Rebuffi, Fong, Ji, and Vedaldi}]{revisiting2020}
Rebuffi, S.-A.; Fong, R.; Ji, X.; and Vedaldi, A. 2020.
\newblock There and Back Again: Revisiting Backpropagation Saliency Methods.
\newblock In \emph{Proceedings of the IEEE/CVF Conference on Computer Vision
  and Pattern Recognition}, 8839--8848.

\bibitem[{Ribeiro, Singh, and Guestrin(2016)}]{Ribiero2016LIME}
Ribeiro, M.~T.; Singh, S.; and Guestrin, C. 2016.
\newblock ``Why Should {I} Trust You?": Explaining the Predictions of Any
  Classifier.
\newblock In \emph{Proceedings of the 22nd {ACM} {SIGKDD} International
  Conference on Knowledge Discovery and Data Mining, San Francisco, CA, USA,
  August 13-17, 2016}, 1135--1144.

\bibitem[{Sandler et~al.(2018)Sandler, Howard, Zhu, Zhmoginov, and
  Chen}]{MobileNetv2}
Sandler, M.; Howard, A.; Zhu, M.; Zhmoginov, A.; and Chen, L.-C. 2018.
\newblock Mobilenetv2: Inverted residuals and linear bottlenecks.
\newblock In \emph{Proceedings of the IEEE conference on computer vision and
  pattern recognition}, 4510--4520.

\bibitem[{Schulz et~al.(2020)Schulz, Sixt, Tombari, and
  Landgraf}]{Schulz2020Restricting}
Schulz, K.; Sixt, L.; Tombari, F.; and Landgraf, T. 2020.
\newblock Restricting the Flow: Information Bottlenecks for Attribution.
\newblock In \emph{International Conference on Learning Representations}.
\newblock \urlprefix\url{https://openreview.net/forum?id=S1xWh1rYwB}.

\bibitem[{Selvaraju et~al.(2017)Selvaraju, Cogswell, Das, Vedantam, Parikh, and
  Batra}]{selvaraju2017grad}
Selvaraju, R.~R.; Cogswell, M.; Das, A.; Vedantam, R.; Parikh, D.; and Batra,
  D. 2017.
\newblock Grad-CAM: Visual Explanations From Deep Networks via Gradient-Based
  Localization.
\newblock In \emph{Proceedings of the IEEE International Conference on Computer
  Vision (ICCV)}.

\bibitem[{Shen, Ma, and Li(2018)}]{shen2018end}
Shen, L.; Ma, Q.; and Li, S. 2018.
\newblock End-to-end time series imputation via residual short paths.
\newblock In \emph{Asian Conference on Machine Learning}, 248--263.

\bibitem[{Shrikumar, Greenside, and Kundaje(2017)}]{shrikumar2017learning}
Shrikumar, A.; Greenside, P.; and Kundaje, A. 2017.
\newblock Learning Important Features Through Propagating Activation
  Differences.
\newblock In Precup, D.; and Teh, Y.~W., eds., \emph{Proceedings of the 34th
  International Conference on Machine Learning}, volume~70 of \emph{Proceedings
  of Machine Learning Research}, 3145--3153. International Convention Centre,
  Sydney, Australia: PMLR.
\newblock \urlprefix\url{http://proceedings.mlr.press/v70/shrikumar17a.html}.

\bibitem[{Simonyan, Vedaldi, and Zisserman(2014)}]{simonyan2013deep}
Simonyan, K.; Vedaldi, A.; and Zisserman, A. 2014.
\newblock Deep Inside Convolutional Networks: Visualising Image Classification
  Models and Saliency Maps.
\newblock In \emph{Workshop at International Conference on Learning
  Representations}.

\bibitem[{Smilkov et~al.(2017)Smilkov, Thorat, Kim, Vi{\'e}gas, and
  Wattenberg}]{smilkov1706smoothgrad}
Smilkov, D.; Thorat, N.; Kim, B.; Vi{\'e}gas, F.; and Wattenberg, M. 2017.
\newblock Smoothgrad: Removing noise by adding noise. arXiv 2017.
\newblock \emph{arXiv preprint arXiv:1706.03825} .

\bibitem[{Srinivas and Fleuret(2019)}]{srinivas2019full}
Srinivas, S.; and Fleuret, F. 2019.
\newblock Full-gradient representation for neural network visualization.
\newblock In \emph{Advances in Neural Information Processing Systems},
  4126--4135.

\bibitem[{Sundararajan, Taly, and Yan(2017)}]{sundararajan2017axiomatic}
Sundararajan, M.; Taly, A.; and Yan, Q. 2017.
\newblock Axiomatic attribution for deep networks.
\newblock In \emph{Proceedings of the 34th International Conference on Machine
  Learning-Volume 70}, 3319--3328. JMLR. org.

\bibitem[{Tan and Le(2019)}]{EfficientNet}
Tan, M.; and Le, Q.~V. 2019.
\newblock EfficientNet: Rethinking Model Scaling for Convolutional Neural
  Networks.
\newblock \emph{CoRR} abs/1905.11946.
\newblock \urlprefix\url{http://arxiv.org/abs/1905.11946}.

\bibitem[{Veit, Wilber, and Belongie(2016)}]{Veit}
Veit, A.; Wilber, M.~J.; and Belongie, S. 2016.
\newblock Residual networks behave like ensembles of relatively shallow
  networks.
\newblock In \emph{Advances in neural information processing systems},
  550--558.

\bibitem[{Wang et~al.(2020)Wang, Wang, Du, Yang, Zhang, Ding, Mardziel, and
  Hu}]{wang2019score}
Wang, H.; Wang, Z.; Du, M.; Yang, F.; Zhang, Z.; Ding, S.; Mardziel, P.; and
  Hu, X. 2020.
\newblock Score-CAM: Score-Weighted Visual Explanations for Convolutional
  Neural Networks.
\newblock In \emph{Proceedings of the IEEE/CVF Conference on Computer Vision
  and Pattern Recognition Workshops}, 24--25.

\bibitem[{Zeiler and Fergus(2014)}]{zeiler2014visualizing}
Zeiler, M.~D.; and Fergus, R. 2014.
\newblock Visualizing and understanding convolutional networks.
\newblock In \emph{European conference on computer vision}, 818--833. Springer.

\bibitem[{Zhang et~al.(2018)Zhang, Bargal, Lin, Brandt, Shen, and
  Sclaroff}]{10.1007/s11263-017-1059-x}
Zhang, J.; Bargal, S.~A.; Lin, Z.; Brandt, J.; Shen, X.; and Sclaroff, S. 2018.
\newblock Top-Down Neural Attention by Excitation Backprop.
\newblock \emph{Int. J. Comput. Vision} 126(10): 1084–1102.
\newblock ISSN 0920-5691.
\newblock \doi{10.1007/s11263-017-1059-x}.
\newblock \urlprefix\url{https://doi.org/10.1007/s11263-017-1059-x}.

\bibitem[{Zhou et~al.(2016)Zhou, Khosla, Lapedriza, Oliva, and
  Torralba}]{zhou2016learning}
Zhou, B.; Khosla, A.; Lapedriza, A.; Oliva, A.; and Torralba, A. 2016.
\newblock Learning deep features for discriminative localization.
\newblock In \emph{Proceedings of the IEEE conference on computer vision and
  pattern recognition}, 2921--2929.

\bibitem[{Zoph and Le(2016)}]{NAS}
Zoph, B.; and Le, Q.~V. 2016.
\newblock Neural architecture search with reinforcement learning.
\newblock \emph{arXiv preprint arXiv:1611.01578} .

\end{thebibliography}
\clearpage

\section{Technical Appendix}
\vspace{0.7cm}

\section{Datasets}
Experiments are conducted on three different datasets: MS COCO 2014 \cite{lin2014microsoft}, PASCAL VOC 2007 \cite{PASCALVOC}, and Severstal \cite{severstalDataset}. The first two datasets are ``natural image" object detection datasets, while the last one is an ``industrial" steel defect detection dataset. They are discussed more in detail in the following subsections.

\subsection{MS COCO 2014 and PASCAL VOC 2007 Datasets}
The MS COCO 2014 dataset features 80 different object classes, each one of a common object. All experimental results are performed on the validation set, which has 40,504 images. The PASCAL VOC 2007 dataset features 20 object classes, and all experimental results for this dataset are performed on its test set, which has 4,952 images. Both datasets are created for object detection and segmentation purposes and contain images with multiple object classes, and images with multiple object instances, making these datasets challenging for XAI algorithms to perform well on.

\subsection{Severstal Dataset}

To extend the analysis of the influence of XAI algorithms beyond natural images, the Severstal steel defect detection dataset was chosen. It was originally hosted on Kaggle as a ``detection" task, which we then converted to a ``classification" task. The original dataset has 12,568 train images under one normal class labeled ``0", and four defective classes numbered 1 through 4. Each image may contain no defect, or one defect, or two and more defects from different classes in it. The ground truth annotations for the segments (masks) are provided in a CSV file, with a single row entry for each class of defect present within each image. The row entries provide the locations of defects, with some entries having several non-contiguous defect locations available.

\begin{figure}[h]
    \centering
    \includegraphics[width=0.95\linewidth]{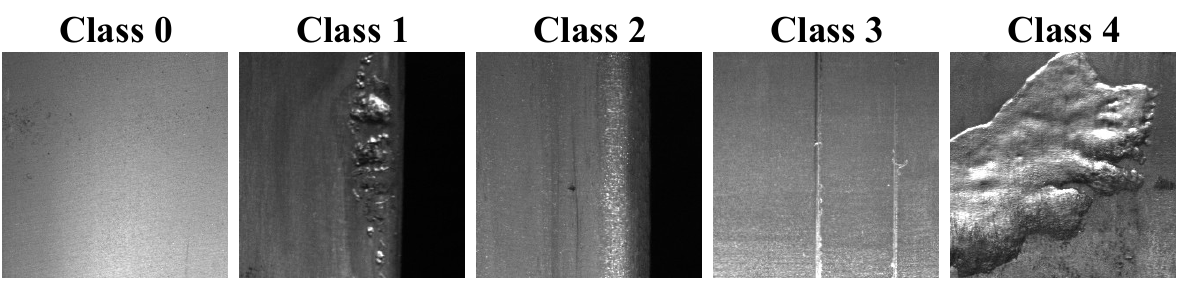} 
    \caption{Sample images with dimension $256\times 256$, from each class of the recast Severstal dataset.}
    \label{fig:severstal_recast}
\end{figure}

The original images were long strips of steel sheets with dimensions 1600 $\times$ 256 pixels. To convert the dataset for our purpose, every training image was cropped (without any overlap) with an initial offset of 32 pixels into 6 individual images of dimensions 256 $\times$ 256 pixels. The few empty (black) images that tended to be located along the sides of the original long strip images were discarded, along with images that had multiple types of defects. This re-formulation left a highly-imbalanced dataset with 5 distinct classes - 0, 1, 2, 3, and 4. Class 0 contains images with no defects, whereas the other four classes have images with only that specific defect group. Fig. \ref{fig:severstal_recast} shows sample images from each class of the recast dataset. The image per class distribution is provided in Table \ref{tab:Severstal_data_distribution}. The training split is 70\% of the data, and the test is the remaining 30\%. From the training data, 20\% is used for validation. The experimental results and qualitative figures of the Severstal dataset are conducted on a subset of the test set using all of the images from classes 1, 2, and 4, and using 500 images from class 3.

\begin{table}[t]
    \centering
    \begin{tabular}{c c c c}
         \multicolumn{4}{c}{\textbf{Severstal: Steel Defect Detection}} \\ [0.5ex]
         \toprule
         \multirow{2}{*}{\textbf{Class}} & \textbf{Training} & \multirow{2}{*}{\textbf{Test set}} & \multirow{2}{*}{\textbf{Total}} \\ 
         & \textbf{set} & & 
         \\ 
         \midrule
         0 & 16620 & 7124 & 23744 \\
         1 & 935 & 401 & 1336 \\
         2 & 147 & 63 & 210 \\
         3 & 8166 & 3500 & 11666 \\
         4 & 971 & 417 & 1388 \\ [0.5ex]
         \bottomrule
    \end{tabular}
    \caption{Data distribution on each class of the recast Severstal dataset, outlining the high data-imbalance among them.}
    \label{tab:Severstal_data_distribution}
\end{table}

\section{Models} 

\begin{table*}[t]
 \centering
 \begin{tabular}{c c c c c c c c c c}
 \toprule[1.2pt]
 \multirow{2}{*}{\textbf{Model}} & \multirow{2}{*}{\textbf{Metric}} &
 \multirow{2}{*}{Grad-CAM} & Grad- & Extremal & \multirow{2}{*}{RISE} & Score- & Integrated & \multirow{2}{*}{FullGrad} &  \multirow{2}{*}{\textbf{SISE}} \\
 & & & CAM++ & Perturbation &  & CAM & Gradient & &  \\ [0.5ex]
 \midrule[1.2pt]
 \multirow{5}{*}{\textbf{VGG16}} & \textbf{EBPG} &
 23.77 & 18.11 & \underline{25.71} & 11.5 & 12.59 & 14.01 & 13.96 & \textbf{28.16} \\
  & \textbf{mIoU} &
 15.04 & \textbf{15.69} & 12.81 & 14.94 & 15.52 & 7.13 & 14.25 & \underline{15.57} \\
  & \textbf{Bbox} &
 \underline{28.98} & 20.48 & 24.93 & 28.9 & 27.8 & 14.54 & 27.52 & \textbf{29.63} \\
 \cmidrule(lr){2-10}
 & \textbf{Drop\%} & 44.46 & 45.63 & 41.86 & 38.69 & \underline{33.73} & 52.73 & 52.39 & \textbf{32.9}  \\
  & \textbf{Increase\%} & 40.28 & 38.33 & 41.30 & 46.05 & \underline{49.26} & 34.11 & 32.68 &  \textbf{50.56}\\ [0.5ex]
 \midrule[1.2pt]
 \multirow{5}{*}{\textbf{ResNet-50}} & \textbf{EBPG} &
25.3 & 17.81 & \underline{27.54} & 11.35 & 12.6 & 14.41 & 14.39 & \textbf{29.43}\\
  & \textbf{mIoU} &
 \textbf{17.89} & 15.8 & 13.61 & 14.69 & 16.36 & 7.24 & 10.14  & \underline{17.03}\\
  & \textbf{Bbox} &
 \underline{32.39} & 28.28 & 26.98 & 29.43 & 29.27 & 14.54 & 19.32 & \textbf{33.34}\\ 
  \cmidrule(lr){2-10}
  & \textbf{Drop\%} &
 \underline{33.42} & 41.71 & 36.24 & 37.93 & 35.06 & 55.38 & 56.83  & \textbf{31.41}\\
  & \textbf{Increase\%} &
 \underline{48.39} & 40.54 & 45.74 & 45.44 & 47.25 & 32.18 & 29.59 & \textbf{49.76}\\ [0.5ex]
 \bottomrule[1.2pt]
 \end{tabular}
 \caption{Results of ground truth-based and model truth-based metrics for state-of-the-art XAI methods along with SISE (proposed) on two networks (VGG16 and ResNet-50) trained on MS COCO 2014 dataset. For each metric, the best is shown in bold, and the second-best is underlined.  Except for Drop\%, the higher is better for all other metrics.}
 \label{tab: coco_metrics}
\end{table*}

\subsection{VGG16 and ResNet-50} 
The top-1 accuracies of the VGG16 and ResNet-50 models (loaded from the TorchRay library \cite{fong2019understanding}) on the test set of the PASCAL VOC 2007 dataset were 56.56 percent and 57.08 percent respectively out of a maximum top-1 accuracy of 64.88 percent, while the top-5 accuracies were 93.29 percent and 93.09 percent respectively out of a maximum top-5 accuracy of 99.99 percent. The top-1 accuracies of the VGG16 and ResNet-50 on the validation set of the MS COCO 2014 dataset were 29.62 percent and 30.25 percent respectively out of a maximum top-1 accuracy of 34.43 percent, while the top-5 accuracies were 69.01 percent and 70.27 percent respectively out of a maximum top-5 accuracy of 93.28 percent.

\begin{table}[ht]
    \centering
    \begin{tabular}{c c c c c c c}
    \multicolumn{2}{c}{}
                &   \multicolumn{5}{c}{\textbf{Predicted Class}} \\
        &       &  \textbf{0} & \textbf{1} & \textbf{2} & \textbf{3} & \textbf{4} \\
        \cmidrule[1.2pt]{2-7}
    \multirow{5}{*}{\rotatebox[origin=c]{90}{\textbf{Actual Class}}}
        & \textbf{0} & 0.89 & 0.011 & 0.0056 & 0.077 & 0.012 \\ [0.5ex]
        & \textbf{1} & 0.27 & 0.59 & 0.02 & 0.12 & 0.0025 \\ [0.5ex]
        & \textbf{2} & 0.095 & 0.032 & 0.71 & 0.16 & 0 \\ [0.5ex]
        & \textbf{3} & 0.12 & 0.014 & 0.004 & 0.85 & 0.0086 \\ [0.5ex]
        & \textbf{4} & 0.15 & 0.0072 & 0.0024 & 0.16 & 0.67 \\[0.5ex]
        \cmidrule[1.2pt]{2-7}
    \end{tabular}
    \caption{Normalized confusion matrix of ResNet-101 model trained on recast Severstal dataset.}
    \label{svs_cm}
\end{table}

\subsection{ResNet-101}
A ResNet-101 model was trained on the recast Severstal dataset using a \textit{Stochastic Gradient Descent} (SGD) optimizer along with a \textit{categorical cross-entropy} loss function. The model is trained for 40 epochs with an initial learning rate of 0.1, which is dropped by half every 5 epochs. Considering the high data imbalance among the classes, the top-1 accuracy of the ResNet-101 model on the test set of the recast Severstal dataset was 86.58 percent, while the top-3 accuracy was 99.60 percent. Table \ref{svs_cm} shows the normalized confusion matrix of this model.

\section{Evaluation}

\begin{figure*}[ht]
    \centering
    \includegraphics[width=1\linewidth]{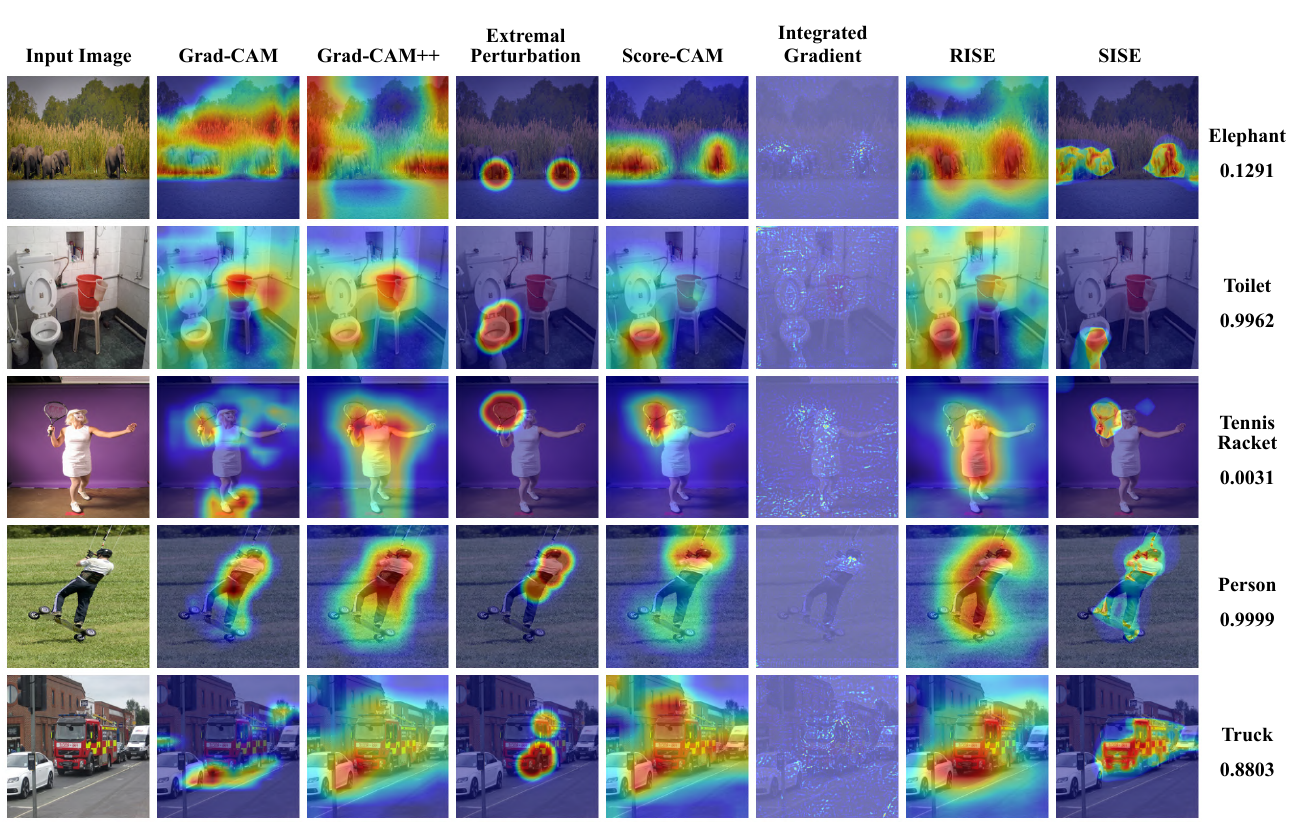}
    \caption{Explanations of SISE along with other conventional methods from a VGG16 model on the MS COCO 2014 dataset.}
    \label{fig:vgg_coco}
\end{figure*}

\begin{figure*}[ht]
    \centering
    \includegraphics[width=1\linewidth]{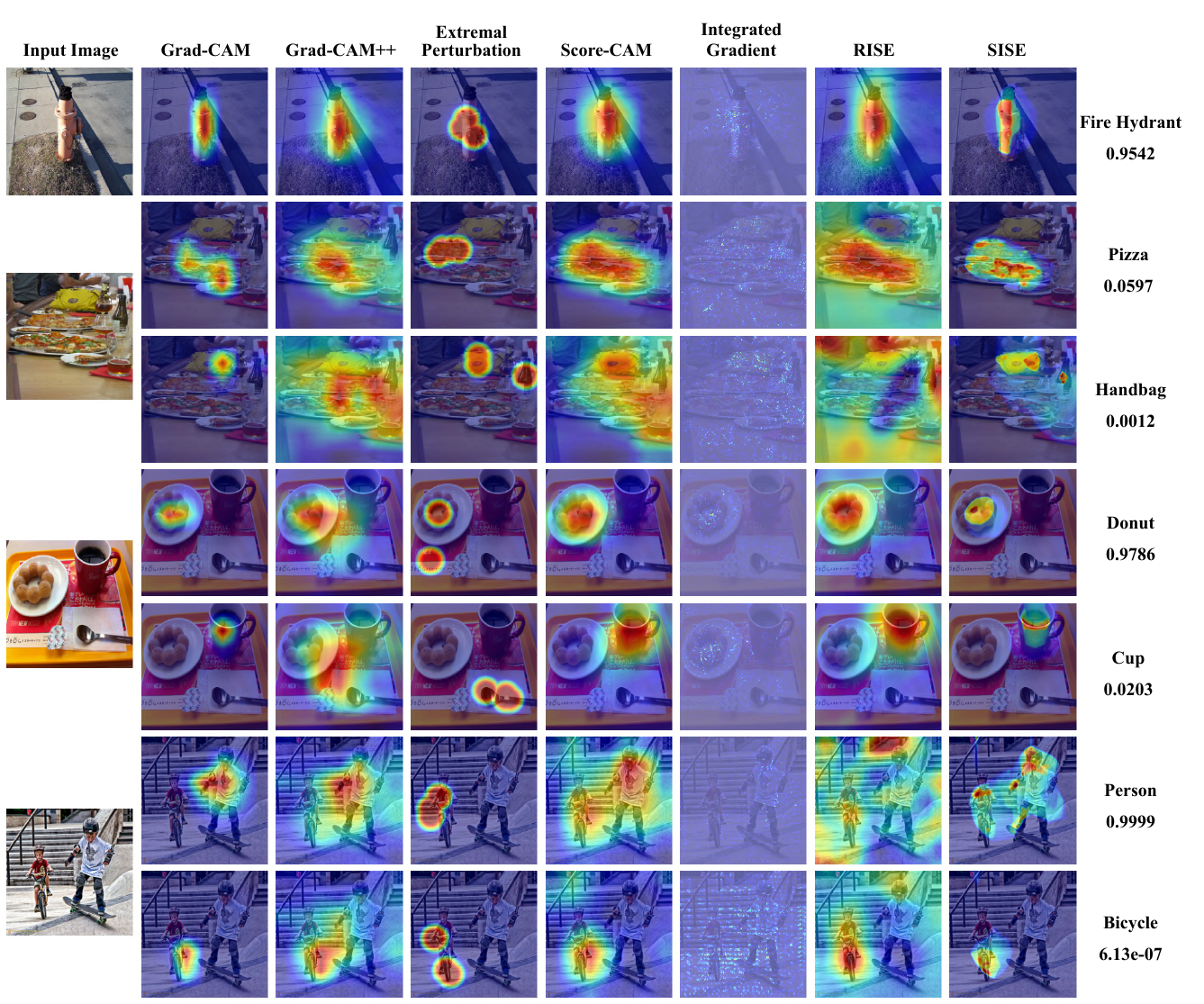}
    \caption{Qualitative results of SISE and other XAI algorithms from the ResNet-50 model trained on the MS COCO 2014 dataset.}
    \label{fig:rs50_coco}
\end{figure*}

In addition to the quantitative evaluation results shared on the main paper, the results of both ground-truth based and model-truth based metrics on the MS COCO 2014 dataset are attached in Table \ref{tab: coco_metrics}. Similar to our earlier results, SISE outperforms other conventional XAI methods in most cases. The MS COCO 2014 data set is more challenging for the explanation algorithms than the PASCAL VOC 2007 dataset because of 
\begin{quote}
    \begin{itemize}
    \item the higher number of object instances
    \item the presence of more extra small objects
    \item the presence of more objects either from the same or different classes in each image (on average)
    \item the lower classification accuracy of the models trained on it (as provided in TorchRay library).
\end{itemize}
\end{quote}
However, the results depicted in Table \ref{tab: coco_metrics} and Figs. \ref{fig:vgg_coco} and \ref{fig:rs50_coco} emphasizes the superior ability of SISE in providing satisfying, high-resolution, and complete explanation maps that provide a precise visual analysis of the model's predictions and perspective.

The benchmark results reported on the Pascal VOC 2007 and MS COCO 2014 datasets are calculated for all ground-truth labels in the test images. For example, if a chosen input image has both ``dog" and ``cat" object instances, then explanations are collected for both { \tt class\_ids} and accounted for in overall performance. SISE's ability to generate class discriminative explanations is represented in this manner. As discussed in the main manuscript, SISE chooses pooling layers to collect feature maps, which are later combined in the fusion module. 
The experiments on the Severstal dataset were performed for only the ground-truth labels, as each test image has exactly one { \tt class\_id} associated with it. 

\begin{figure*}[ht]
    \centering
    \includegraphics[width=0.95\linewidth]{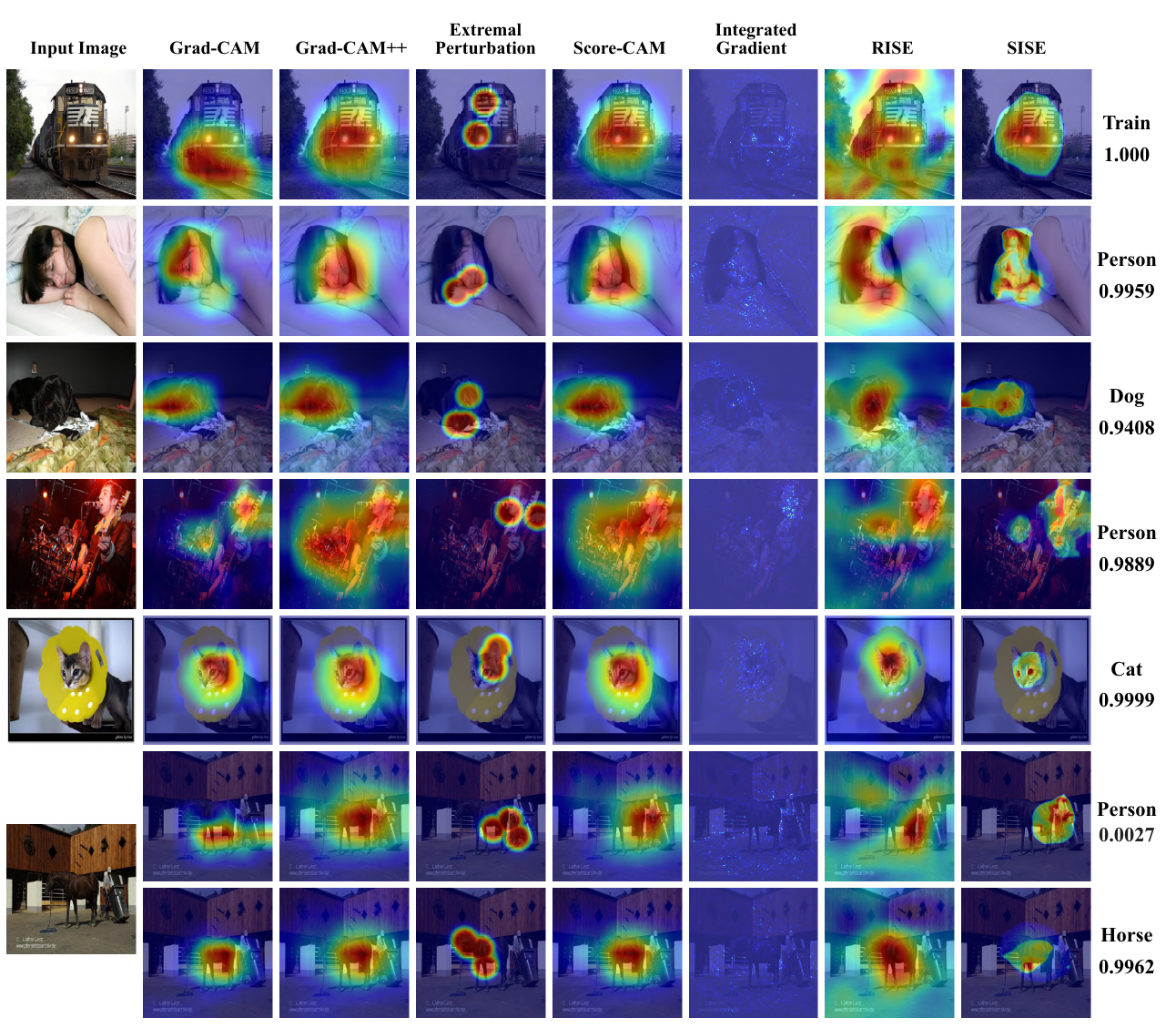} 
    \caption{Qualitative comparison of SISE with other state-of-the-art XAI methods with a ResNet-50 model on the Pascal VOC 2007 dataset.}
    \label{fig:rs50_voc}
\end{figure*}

\begin{figure*}[ht]
    \centering
    \includegraphics[width=0.95\linewidth]{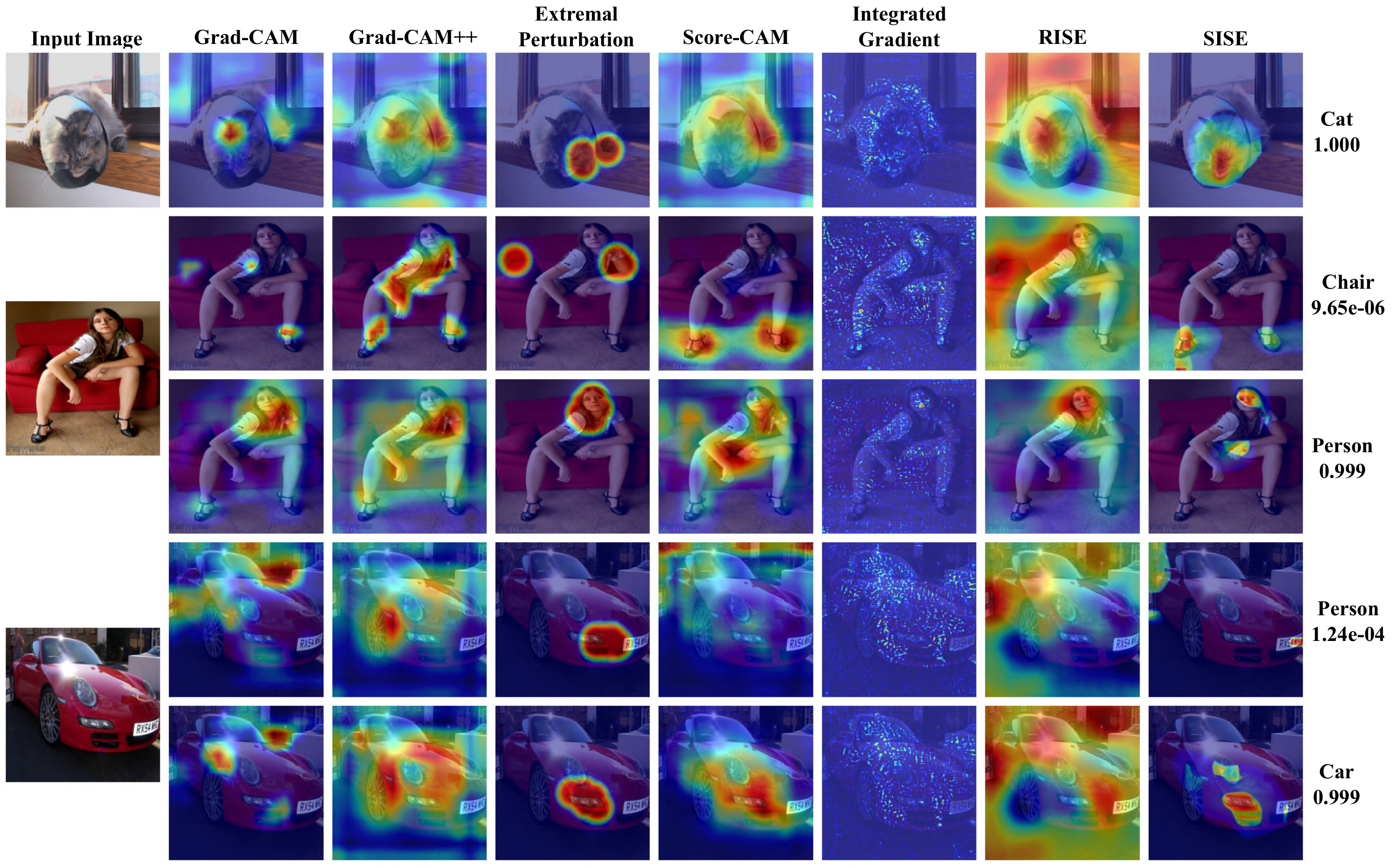} 
    \caption{Comparison of SISE explanations generated with a VGG16 model on the Pascal VOC 2007 dataset.}
    \label{fig:vgg_voc}
\end{figure*}

\begin{figure*}[ht]
    \centering
    \includegraphics[width=0.9\linewidth]{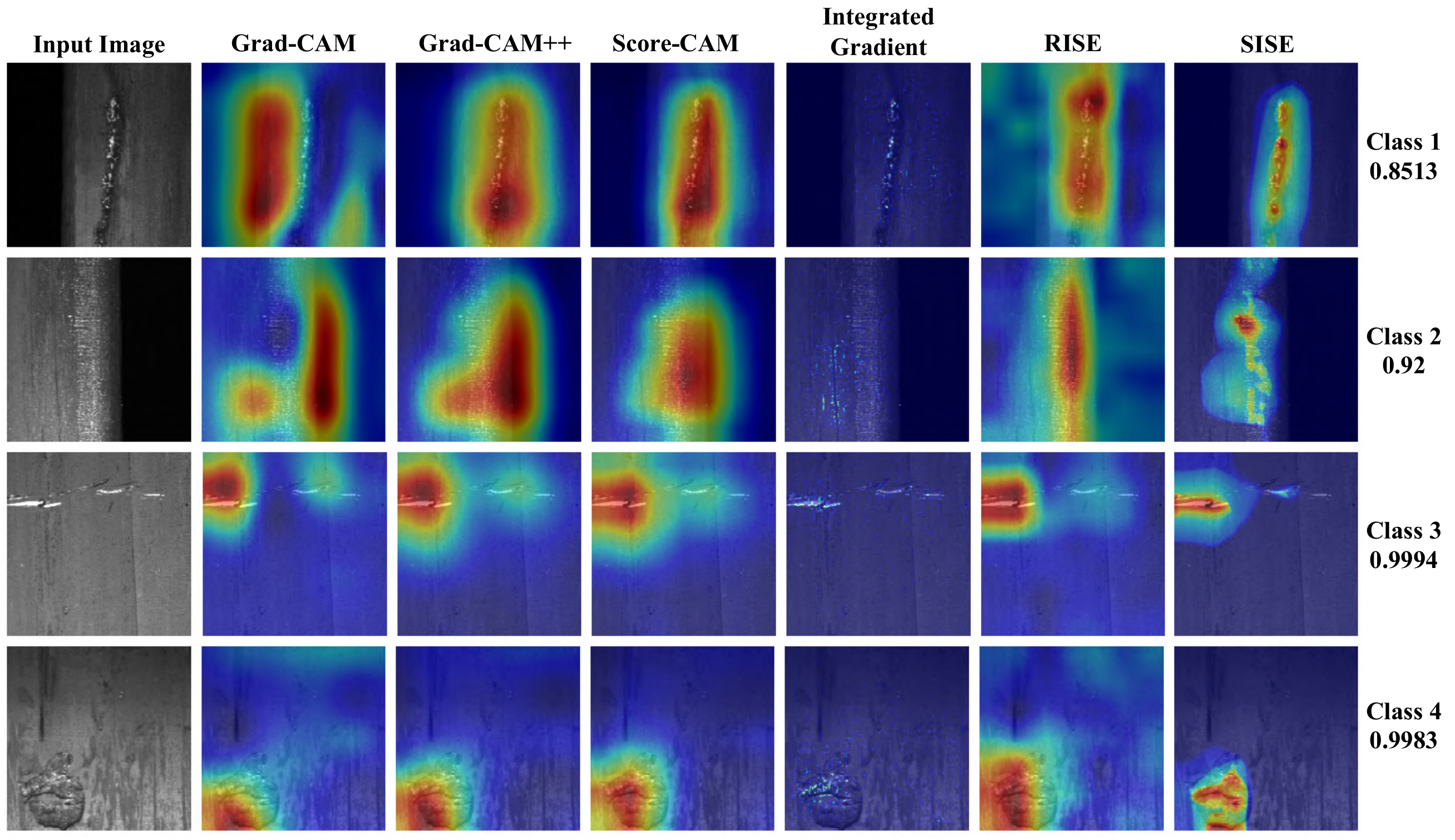} 
    \caption{Qualitative results of SISE and other XAI algorithms from the ResNet-101 model trained on the recast Severstal dataset.}
    \label{fig:rs101_svs}
\end{figure*}

A detailed qualitative analysis of SISE explanations compared with other state-of-the-art XAI algorithms on the discussed models on Pascal VOC 2007 and recast Severstal datasets are shown in Figs. \ref{fig:rs50_voc}, \ref{fig:vgg_voc} and \ref{fig:rs101_svs} respectively. Figs. \ref{fig:vgg_coco} and \ref{fig:rs50_coco} show a similar comparative analysis on MS COCO 2014 dataset.

\begin{figure*}[htbp]
    \centering
    \includegraphics[width=\linewidth]{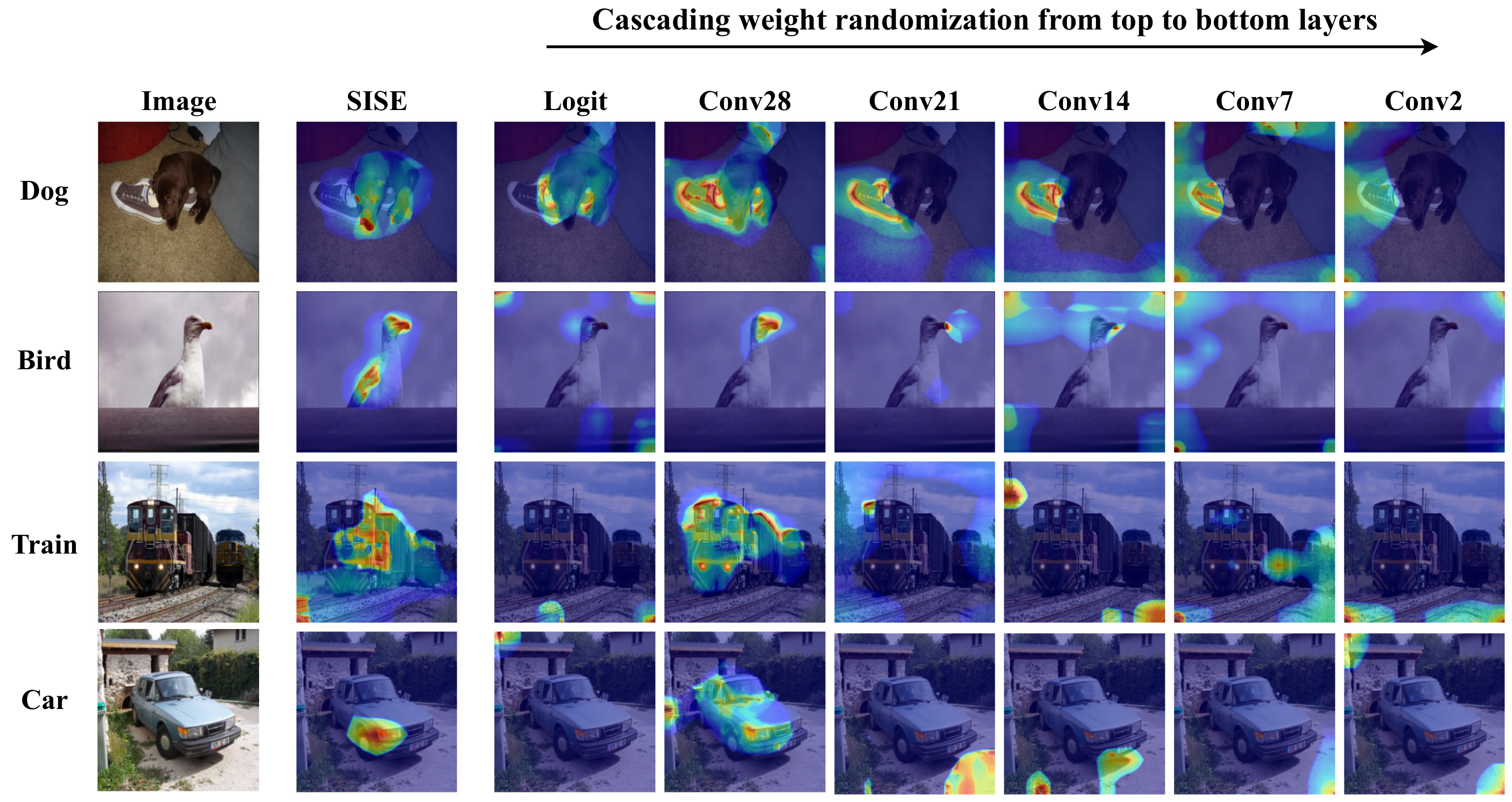} 
    \caption{Sanity check experimentation of SISE as per \cite{KimSanity} by randomizing a VGG16 model's (pre-trained on Pascal VOC 2007 dataset) parameters.}
    \label{fig:sanity_random}
\end{figure*}

\section{Ablation Study}
\label{sec: ablation}

\begin{table}[thbp]
 \centering
 \begin{tabular}{c c c c c}
 \toprule[1.2pt]
 \textbf{Metric} &
 $\mu$ = 0 & $\mu$ = 0.3 & $\mu$ = 0.5 & $\mu$ = 0.75 \\
 \midrule[1.2pt]
 \textbf{EBPG} &
66.08 & 66.54 & 65.84 & 62.5\\
  \textbf{mIoU} &
 31.37 & 31.5 & 30.63 & 28.51 \\
  \textbf{Bbox} &
  61.59 & 61.45 & 59.83 & 56.53 \\
  \cmidrule(lr){1-5}
  \textbf{Drop\%} &
 30.92 & 31.5 & 33.31 & 38.83 \\
  \textbf{Increase\%} &
 40.22 & 40.05 & 38.36 & 36.09 \\ 
 \cmidrule(lr){1-5}
  \textbf{Runtime (s)} &
 9.21 & 2.18 & 0.65 & 0.38 \\
 \bottomrule[1.2pt]
 \end{tabular}
 \caption{Performance and runtime results of SISE with respect to the parameter $\mu$, on a ResNet-50 network trained on PASCAL VOC 2007 dataset. Except for Drop\% and runtime (in seconds), the higher is better for all other metrics.}
 \label{tab: ablation}
\end{table}

As stated in the main manuscript, in the second phase of SISE, each set of feature maps is valuated by backpropagating the signal from the output of the model to the layer from which the feature maps are derived. In this stage, after normalizing the backpropagation-based scores, a threshold $\mu$ is applied to each set, so that the feature maps passing the threshold are converted to attribution masks and utilized in the next steps, while the others are discarded. Some of these feature maps do not contain signals that lead the model to make a firm prediction since they represent the attributions related to the instances of the other classes (rather than the class of interest). These feature maps are expected to be identified by reaching zero or negative backpropagation-based scores. Getting rid of them by setting the threshold parameter $\mu$ to 0 ($\mu$ is defined in the main manuscript) will improve our method, not only by increasing its speed but also by enabling us to analyze the model's decision making process more precisely.

\begin{figure}[thbp]
    \centering
    \includegraphics[width=1\linewidth]{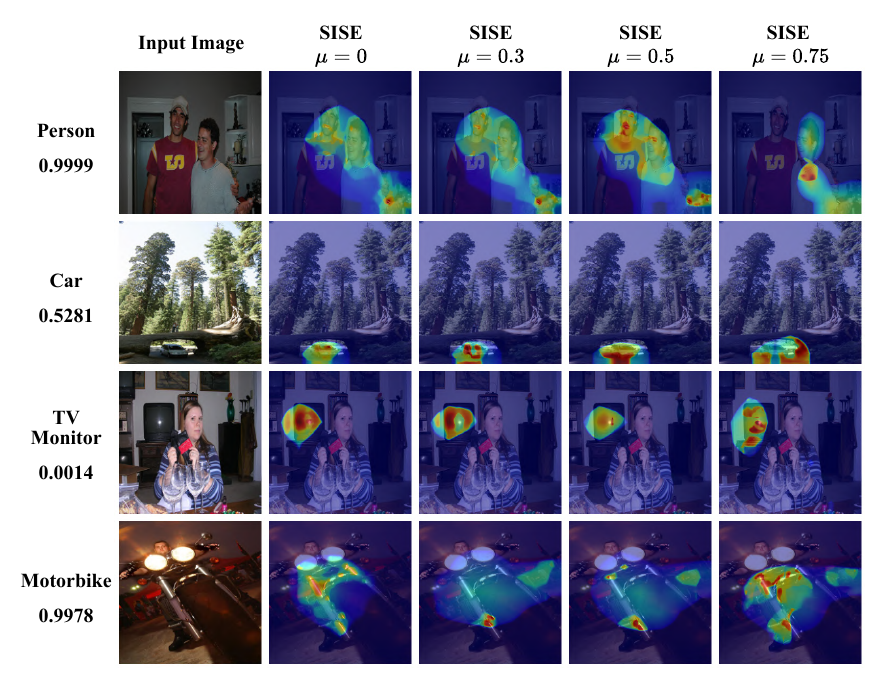} 
    \caption{Effect of SISE's $\mu$ variation on a ResNet-50 model trained on Pascal VOC 2007 dataset.}
    \label{fig:mu_effect}
\end{figure}

By increasing the threshold parameter $\mu$, a trade-off between performance and speed is reached. When this parameter is slightly increased, SISE will discard feature maps with low positive backpropagation-based scores, which is expected not to make a considerable impact on the output explanation map. The higher the parameter $\mu$ is though, the more deterministic feature maps are discarded, causing more degradation in SISE's performance.

To verify these interpretations, we have conducted an ablation analysis on the PASCAL VOC 2007 test set. As stated in the main manuscript, the model truth-based metrics (Drop\% and Increase\%) are the most important metrics revealing the sensitivity of SISE's performance with respect to its threshold parameter. According to our results as depicted in Table \ref{tab: ablation} and Fig. \ref{fig:mu_effect}, the ground truth-based results also follow approximately the same trend for the effect of $\mu$ variation. Consequently, our results show that by adjusting this hyper-parameter, a dramatic increase in SISE's speed is gained in turn with a slight compromise in its explanation ability. 

Since the behavior of our method concerning this hyper-parameter does not depend on the model and the dataset employed, it can be consistently fine-tuned, based on the requirements of the end-user. 

\section{Sanity Check} 

\begin{figure}[th]
    \centering
    \includegraphics[width=0.9\linewidth]{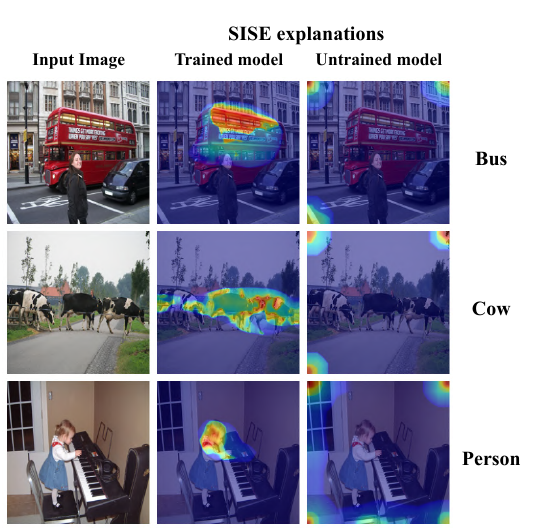} 
    \caption{SISE results from a VGG16 model trained on Pascal VOC 2007 dataset with an untrained VGG16 model.}
    \label{fig:sanity_train_untrain}
\end{figure}

In addition to the comprehensive quantitative experiments presented in the main manuscript and this appendix, we also verified the sensitivity of our explanation algorithm to the model's parameters, illustrating that our method adequately explains the relationship between the input and output that the model reaches. As introduced by \cite{KimSanity}, sanity checks on explanation methods can be conducted either by randomizing the model's parameters or retraining the model with the same training data, but with random labels. In this work, we performed sanity checks on our method by randomizing the parameters of the model. To do so, we have randomized the weight and bias parameters on the VGG16 trained on PASCAL VOC 2007 dataset provided by \cite{fong2019understanding}. Fig. \ref{fig:sanity_random} represents the results of sanity checks for some input images. The layers for which the parameters to be randomized are selected in a top to bottom manner, as specified in the figure. Each row shows the effect on the output explanation maps for an image when we perturb the parameters in more layers.  According to this figure, SISE shows alterations in explanation maps, while dealing with highly perturbed models. Hence, SISE passes our sanity check.



To access SISE's explanation beyond a few evaluation metrics, another sanity check was performed. Fig. \ref{fig:sanity_train_untrain} attached shows such experimentation where an untrained VGG16 model was directly compared with our Pascal VOC dataset trained VGG16 model. SISE doesn't generate quality explanations from the untrained model, insisting that our method not just provide ``featured regions" obtained through convolutional operations, but depict the actual ``attributed regions" affecting the model's decision.

\section{Complexity Evaluation}

\begin{table}[th]
 \centering
 \begin{tabular}{c c c}
 \toprule[1.2pt]
 \multirow{2}{*}{\textbf{XAI Method}} & \textbf{Runtime on}  & \textbf{Runtime on}\\
 & \textbf{VGG16 (s)} & \textbf{ResNet-50 (s)} \\ 
 \midrule[1.2pt]
 Grad-CAM & \textbf{0.006} & \textbf{0.019}\\
 Grad-CAM++ & \textbf{0.006} & \underline{0.020}\\
 Extremal Perturbation & 87.42 & 78.37\\
 RISE & 64.28 & 26.08\\
 Score-CAM & 5.90 & 18.17\\
 Integrated Gradient & \underline{0.68} & 0.52\\
 FullGrad & 18.69 & 34.03\\
 \textbf{SISE} & 5.90 & 9.21\\
 \bottomrule[1.2pt]
 \end{tabular}
 \caption{Results of runtime evaluation of SISE along with other algorithms on a Tesla T4 GPU with 16GB of memory.}
 \label{tab: sev_runtime}
\end{table}

A runtime test was conducted to compare the complexity of the different XAI methods with SISE, timing how long it took for each algorithm to generate an explanation map. It was performed with a Tesla T4 GPU with 16GB of memory on both a VGG16 and ResNet-50 model and attached as Table \ref{tab: sev_runtime}.

Reported runtimes were averaged over 100 trials using a random image from the PASCAL VOC 2007 test set for each trial. Grad-CAM and Grad-CAM++ are the fastest methods when applied to both models. This is expected as they operate using only one main forward pass and one backward pass. Our method, SISE, is not the fastest, and the main bottleneck in its runtime is the number of feature maps extracted and used from the CNN. This is addressed by adjusting $\mu$, as discussed in the `Ablation Study' section.


\end{document}